\pdfoutput=1

\documentclass[11pt]{article}

\usepackage{acl}

\usepackage{times}
\usepackage{latexsym}
\usepackage{subfig}

\usepackage[T1]{fontenc}

\usepackage[utf8]{inputenc}

\usepackage{microtype}
\usepackage{graphicx}
\usepackage{mathtools}
\usepackage{amsmath}
\usepackage{amsfonts}
\usepackage{makecell}
\usepackage{multirow}
\usepackage{soul}
\usepackage{enumitem}
\usepackage{xcolor,colortbl}
\usepackage{tabto}
\definecolor{LightGreen}{rgb}{0.8,1,0.89}
\definecolor{LightCyan}{rgb}{0.88,1,1}

\newcommand{\dataset}{\textsc{WITS}}

\newcommand{\model}{\textsc{MAF}}
\newcommand{\modelTA}{\textsc{MAF-TA\textsubscript{B}}}
\newcommand{\modelTV}{\textsc{MAF-TV\textsubscript{B}}}
\newcommand{\modelTAV}{\textsc{MAF-TAV\textsubscript{B}}}

\newcommand{\modelTAm}{\textsc{MAF-TA\textsubscript{m}}}
\newcommand{\modelTVm}{\textsc{MAF-TV\textsubscript{m}}}
\newcommand{\modelTAVm}{\textsc{MAF-TAV\textsubscript{m}}}

\newcommand{\task}{\textsc{SED}}

\newcommand\blfootnote[1]{%
  \begingroup
  \renewcommand\thefootnote{}\footnote{#1}%
  \addtocounter{footnote}{-1}%
  \endgroup
}

\title{\textit{When did you become so smart, oh wise one?!} Sarcasm Explanation in Multi-modal Multi-party Dialogues}

\author{
    Shivani Kumar$^{*}$, Atharva Kulkarni$^{*}$, Md Shad Akhtar, Tanmoy Chakraborty \\
    \textit{Indraprastha Institute of Information Technology Delhi, India} \\
    \texttt{\{shivaniku, atharvak, shad.akhtar, tanmoy\}@iiitd.ac.in}
}

\begin{document}
\maketitle

\begin{abstract}
Indirect speech such as sarcasm achieves a constellation of discourse goals in human communication. While the indirectness of figurative language warrants speakers to achieve certain pragmatic goals, it is challenging for AI agents to comprehend such idiosyncrasies of human communication. Though sarcasm identification has been a well-explored topic in dialogue analysis, for conversational systems to truly grasp a conversation's innate meaning and generate appropriate responses, simply detecting sarcasm is not enough; it is vital to explain its underlying sarcastic connotation to capture its true essence. In this work, we study the discourse structure of sarcastic conversations and propose a novel task -- {\bf \em Sarcasm Explanation in Dialogue (\task)}. Set in a multimodal and code-mixed setting, the task aims to generate natural language explanations of satirical conversations. To this end, we curate \textbf{\dataset}, a new dataset to support our task. We propose \textbf{\model}\ (Modality Aware Fusion), a multimodal context-aware attention and global information fusion module to capture multimodality and use it to benchmark \dataset. The proposed attention module surpasses the traditional multimodal fusion baselines and reports the best performance on almost all metrics. Lastly, we carry out detailed analyses both quantitatively and qualitatively.

\end{abstract}

\vspace{-1em}
\begin{figure}
    \centering
    \includegraphics[width=\columnwidth]{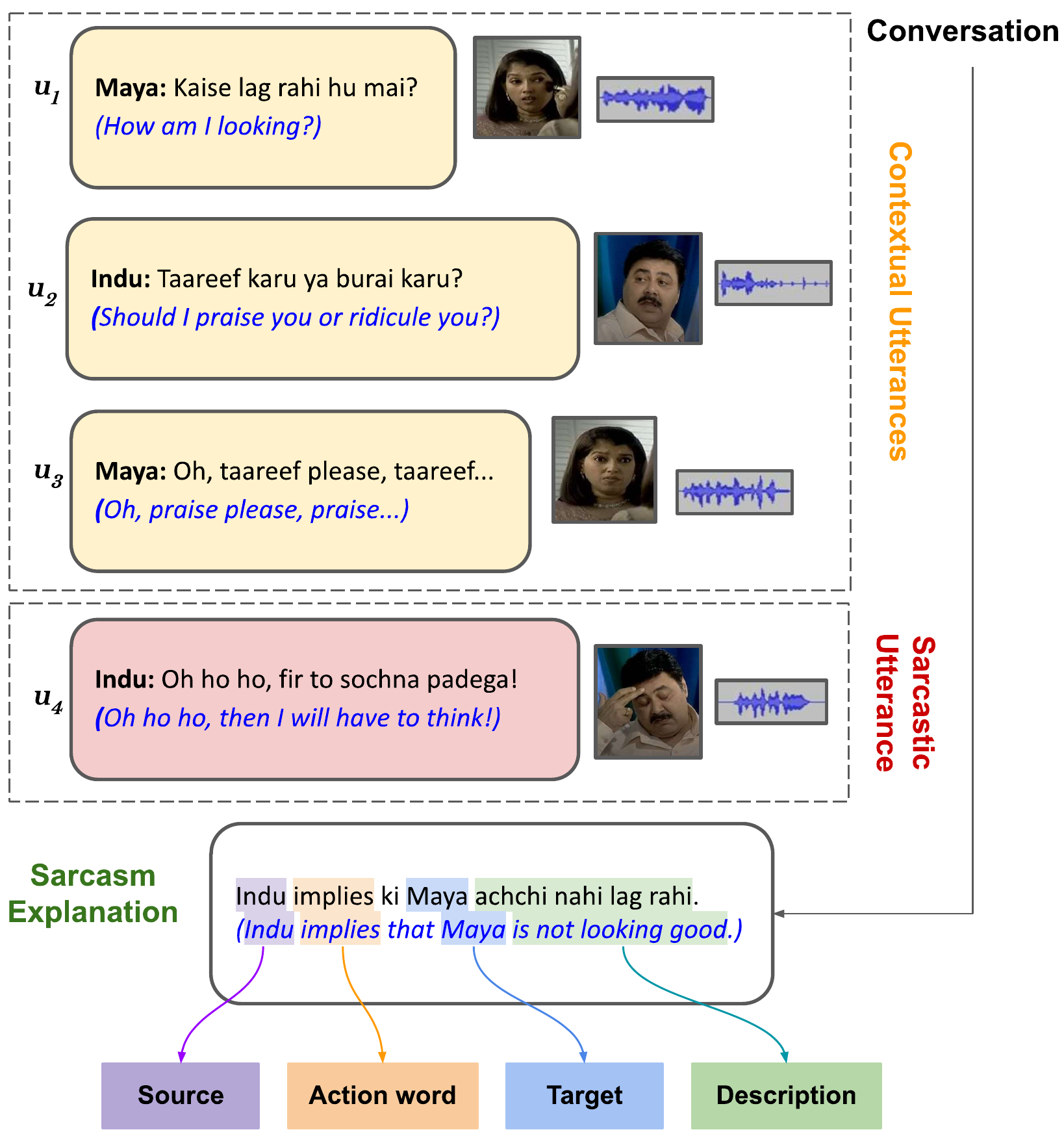}
    \caption{Sarcasm Explanation in Dialogues (\task). Given a sarcastic dialogue, the aim is to generate a natural language explanation for the sarcasm in it. \textit{\color{blue}Blue text represents the English translation for the text.}}
    \label{fig:se_eg}
    \vspace{-5mm}
\end{figure}

\section{Introduction}

The\blfootnote{$^*$Equal contribution} use of figurative language serves many communicative purposes and is a regular feature of both oral and written communication \citep{roberts1994people}. 
Predominantly used to induce humour, criticism, or mockery \citep{colston1997salting}, paradoxical language is also used in concurrence with hyperbole to show surprise \citep{colston1998you} as well as highlight the disparity between expectations and reality \citep{ivanko2003context}. While the use and comprehension of sarcasm is a cognitively taxing process \citep{olkoniemi2016individual}, psychological evidence advocate that it positively correlates with the receiver's theory of mind (ToM) \citep{wellman2014making}, i.e., the capability to interpret and understand another person's state of mind. Thus, for NLP systems to emulate such anthropomorphic intelligent behavior, they must not only be potent enough to identify sarcasm but also possess the ability to comprehend it in its entirety. To this end, moving forward from sarcasm identification, we propose the novel task of \textit{\textbf{Sarcasm Explanation in Dialogue}} ({\bf \task}).

For dialogue agents, understanding sarcasm is even more crucial as there is a need to normalize its sarcastic undertone and deliver appropriate responses. Conversations interspersed with sarcastic statements often use contrastive language to convey the opposite of what is being said. In a real-world setting, understanding sarcasm goes beyond negating a dialogue's language and involves the acute comprehension of audio-visual cues. Additionally, due to the presence of essential temporal, contextual, and speaker-dependent information, sarcasm understanding in conversation manifests as a challenging problem. Consequently, many studies in the domain of dialogue systems have investigated sarcasm from textual, multimodal, and conversational standpoints \cite{10.1162/coli_a_00336, castro2019multimodal, oraby2017serious, 9442359}. However, baring some exceptions \citep{mishra-etal-2019-modular, dubey2019deep, chakrabarty-etal-2020-r}, research on figurative language has focused predominantly on its identification rather than its comprehension and normalization. This paper addresses this gap by attempting to generate natural language explanations of satirical dialogues.

To illustrate the proposed problem statement, we show an example in Figure \ref{fig:se_eg}. It contains a dyadic conversation of four utterances $\langle u_1, u_2, u_3, u_4\rangle$, where the last utterance ($u_4$) is a sarcastic remark. Note that in this example, although the opposite of what is being said is, \textit{``I don't have to think about it,"} it is not what the speaker means; thus, it enforces our hypothesis that sarcasm explanation goes beyond simply negating the dialogue's language. The discourse is also accompanied by ancillary audio-visual markers of satire such as an ironical intonation of the pitch, a blank face, or roll of the eyes. Thus, conglomerating the conversation history, multimodal signals, and speaker information,  \task\ aims to generate a coherent and cohesive natural language explanation associated with sarcastic dialogues.

For the task at hand, we extend \textsc{MaSaC} \cite{9442359} -- a sarcasm detection dataset for code-mixed conversations -- by augmenting it with natural language explanations for each sarcastic dialogue. We name the dataset \dataset\footnote{\dataset: ``Why Is This Sarcastic"}. The dataset is a compilation of sarcastic dialogues from a popular Indian TV show. Along with the textual transcripts of the conversations, the dataset also contains multimodal signals of audio and video.

We experiment with unimodal as well as multimodal models to benchmark \dataset. Text, being the driving force of the explanations, is given the primary importance, and thus, we compare a number of established text-based sequence-to-sequence systems on \dataset. To incorporate multimodal information, we propose a unique fusion scheme of \textit{Multimodal Context-Aware Attention} (MCA2). Inspired by  \citet{Yang_Li_Wong_Chao_Wang_Tu_2019}, this attention variant facilitates deep semantic interaction between the multimodal signals and textual representations by conditioning the key and value vectors with audio-visual information and then performing dot product attention with these modified vectors. The generated audio and video information-informed textual representations are then combined using the \textit{Global Information Fusion Mechanism} (GIF). The gating mechanism of {GIF} allows for the selective inclusion of information relevant to the satirical language and also prohibits any multimodal noise from seeping into the model. We further propose \model\ (\textit{Modality Aware Fusion}) module where the aforementioned mechanisms are introduced in the Generative Pretrained Models (GPLMs) as adapter modules. Our fusion strategy outperforms the text-based baselines and the traditional multimodal fusion schemes in terms of multiple text-generation metrics. Finally, we conduct a comprehensive quantitative and qualitative analysis of the generated explanations.

In a nutshell, our contributions are four fold:
\begin{itemize}[leftmargin=*,topsep=0pt]
\setlength{\itemsep}{0pt}
\setlength{\parskip}{0pt}
\setlength{\parsep}{0pt}
    \item We propose Sarcasm Explanation in Dialogue (\task), \textbf{a novel task} aimed at generating a natural language explanation for a given sarcastic dialogue, elucidating the intended irony.
    \item We extend an existing sarcastic dialogue dataset, to curate \dataset, \textbf{a novel dataset} containing human annotated gold standard explanations.
    \item We \textbf{benchmark our dataset} using \modelTAV\ and \modelTAVm\, variants of BART and mBART, respectively, that incorporate the audio-visual cues using a unique context-aware attention mechanism.
    \item We carry out \textbf{extensive quantitative and qualitative analysis} along with human evaluation to assess the quality of the generated explanations.
\end{itemize}
\paragraph{Reproducibility:} The source codes and the dataset can be found here: \href{https://github.com/LCS2-IIITD/MAF.git}{https://github.com/LCS2-IIITD/MAF.git}.

\section{Related Work}
\paragraph{Sarcasm and Text:}
\citet{1joshi2017automatic} presented a well-compiled survey on computational sarcasm where the authors expanded on the relevant datasets, trends, and issues for automatic sarcasm identification. Early work in sarcasm detection dealt with standalone text inputs like tweets and reviews \cite{kreuz-caucci:2007:sarcasm:lexical, tsur:sarcasm:2010, joshi:sarcsam:incongruity:2015, peled-reichart-2017-sarcasm}. These initial works mostly focused on the use of linguistic and lexical features to spot the markers of sarcasm \citep{kreuz-caucci:2007:sarcasm:lexical,tsur:sarcasm:2010}. 
More recently, attention-based architectures are  proposed to harness the inter- and intra-sentence relationships in texts for efficient sarcasm identification \citep{tay-etal-2018-reasoning, xiong2019sarcasm, srivastava-etal-2020-novel}. Analysis of figurative language has also been extensively explored in conversational AI setting. \citet{ghosh2017role} utilised attention-based RNNs to identify sarcasm in the presence of context. Two separate LSTMs-with-attention were trained for the two inputs (sentence and context) and their hidden representations were combined during the prediction.

The study of sarcasm identification has also expanded beyond the English language. \citet{bharti2017harnessing} collected a Hindi corpus of $2000$ sarcastic tweets and employed rule-based approaches to detect sarcasm. \citet{swami2018corpus} curated a dataset of $5000$ satirical Hindi-English code-mixed tweets and used n-gram feature vectors with various ML models for sarcasm detection. Other notable studies include Arabic \citep{abu-farha-magdy-2020-arabic}, Spanish \citep{ortega2019overview}, and Italian \citep{cignarella2018overview} languages.

\paragraph{Sarcasm and Multimodality:} 
In the conversational setting, MUStARD, a multimodal, multi-speaker dataset compiled by \citet{castro2019multimodal} is considered the benchmark for multimodal sarcasm identification. \citet{chauhan-etal-2020-sentiment} leveraged the intrinsic interdependency between emotions and sarcasm and devised a multi-task framework for multimodal sarcasm detection. Currently, \citet{Hasan_Lee_Rahman_Zadeh_Mihalcea_Morency_Hoque_2021} performed the best on this dataset with their humour knowledge enriched transformer model. Recently, \citet{9442359} proposed a code-mixed multi-party dialogue dataset, \textsc{MaSaC}, for sarcasm and humor detection. In the bimodal setting, sarcasm identification with tweets containing images has also been well explored \citep{cai-etal-2019-multi, xu-etal-2020-reasoning, pan-etal-2020-modeling} .

\paragraph{Beyond Sarcasm Identification:} While studies in computational sarcasm have predominantly focused on sarcasm identification, some forays have been made into other domains of figurative language analysis. \citet{dubey2019deep} initiated the work of converting sarcastic utterances into their non-sarcastic interpretations using deep learning. In another direction, \citet{mishra-etal-2019-modular} devised a modular unsupervised technique for sarcasm generation by introducing context incongruity through fact removal and incongruous phrase insertion. Following this, \citet{chakrabarty-etal-2020-r} proposed a retrieve-and-edit-based unsupervised framework for sarcasm generation. Their proposed model leverages the valence reversal and semantic incongruity to generate sarcastic sentences from their non-sarcastic counterparts. 

In summary, much work has been done in sarcasm detection, but little, if any, effort has been placed into {\em explaining the irony behind sarcasm}. This paper attempts to fill this gap by proposing a new problem definition and a supporting dataset.

\begin{table}[t]
\centering
        \resizebox{\columnwidth}{!}{%
            \begin{tabular}{|p{5em}|p{5em}|p{5em}|p{5em}|}
                \hline
                 \textbf{\# Dlgs} & \textbf{\# Utts} & \textbf{\# Eng utts} & \textbf{\# Hin utts} \\ \hline
                2240 & 9080 & 101 & 1453 \\ \hline \hline
                \textbf{\# CM utts} & \textbf{Avg. utt/dlg} & \textbf{Avg. sp/dlg} & \textbf{Avg. words/utt} \\ \hline
                7526 & 4.05 & 2.35 & 14.39 \\ \hline \hline
                \textbf{Avg. words/dlg} & \textbf{Vocab size} & \textbf{Eng vocab size} & \textbf{Hin vocab size} \\ \hline
                58.33 & 10380 & 2477 & 7903 \\ \hline
            \end{tabular}
        }\\ \vspace{2mm}
\caption{Statistics of dialogs present in \dataset.}
\label{tab:data_stats}
\vspace{-1em}
\end{table}
\begin{figure*}[ht]
\resizebox{\textwidth}{!}{
\centering
    \resizebox{0.33\textwidth}{!}{%
    \subfloat[Utterance length distribution]{
        \includegraphics[width=0.33\textwidth]{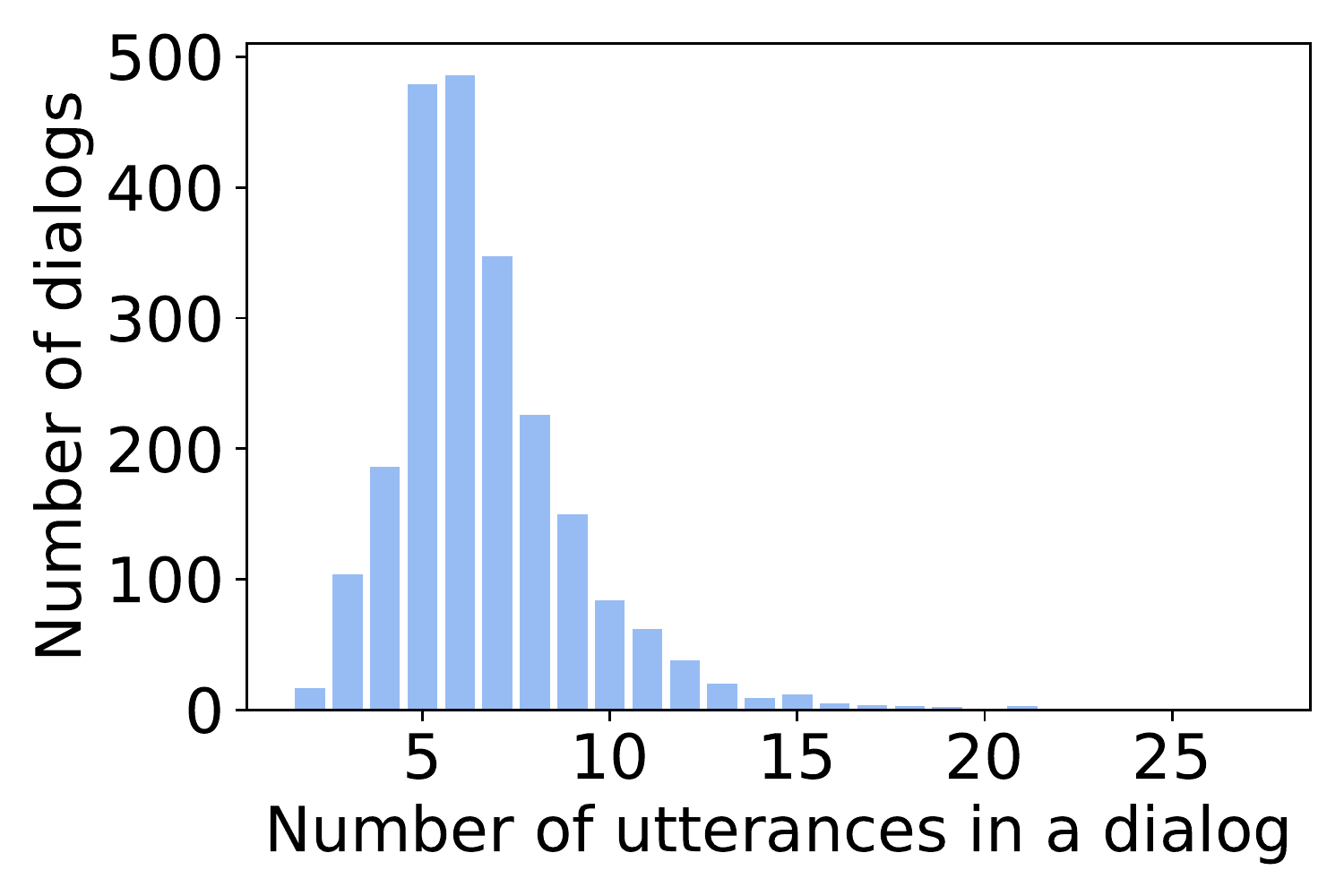}
    }}
    \resizebox{0.33\textwidth}{!}{%
    \subfloat[Speaker distribution]{
        \includegraphics[width=0.33\textwidth]{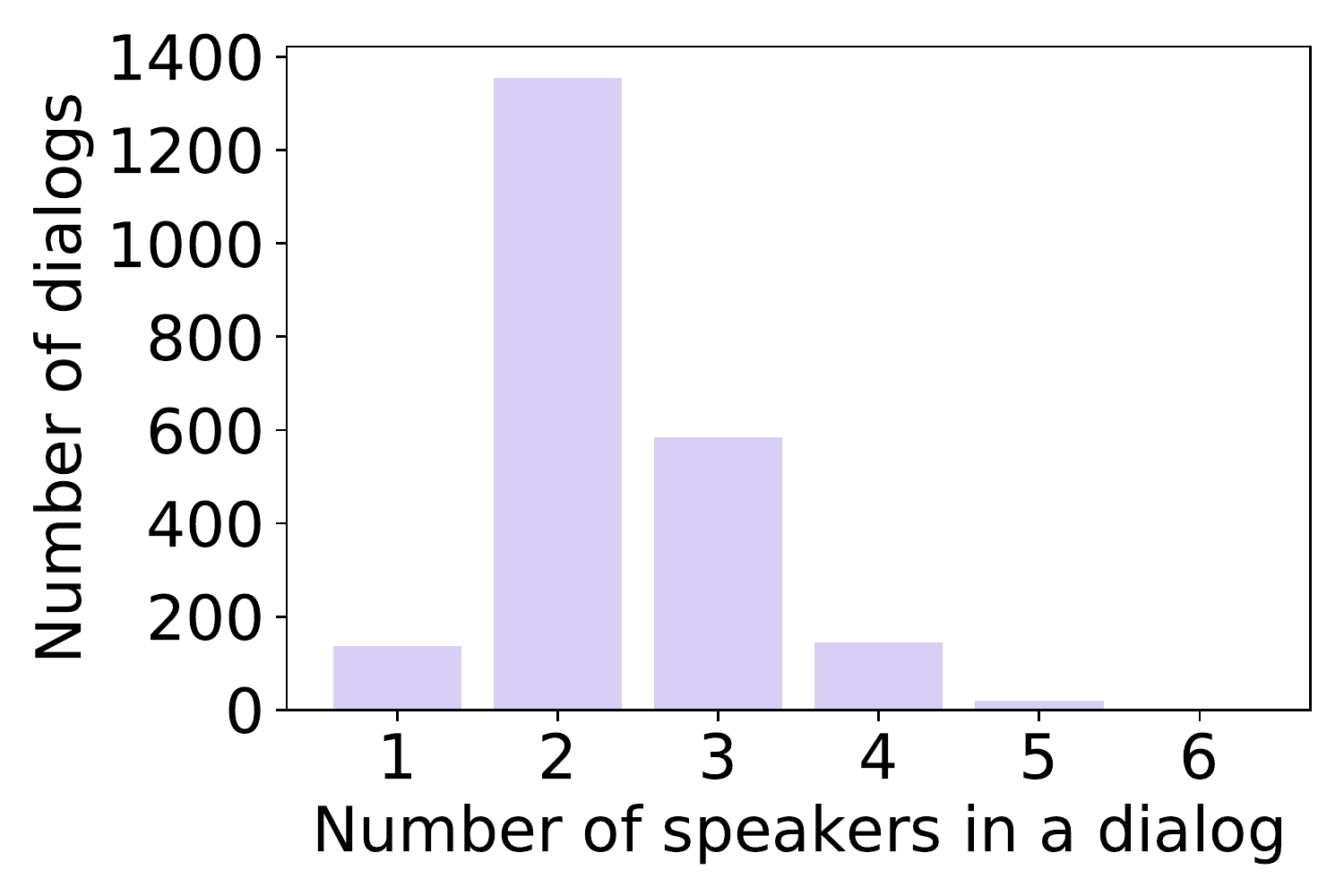}
    }}
    \resizebox{0.33\textwidth}{!}{%
    \subfloat[Source-target pair distribution]{
        \includegraphics[width=0.33\textwidth]{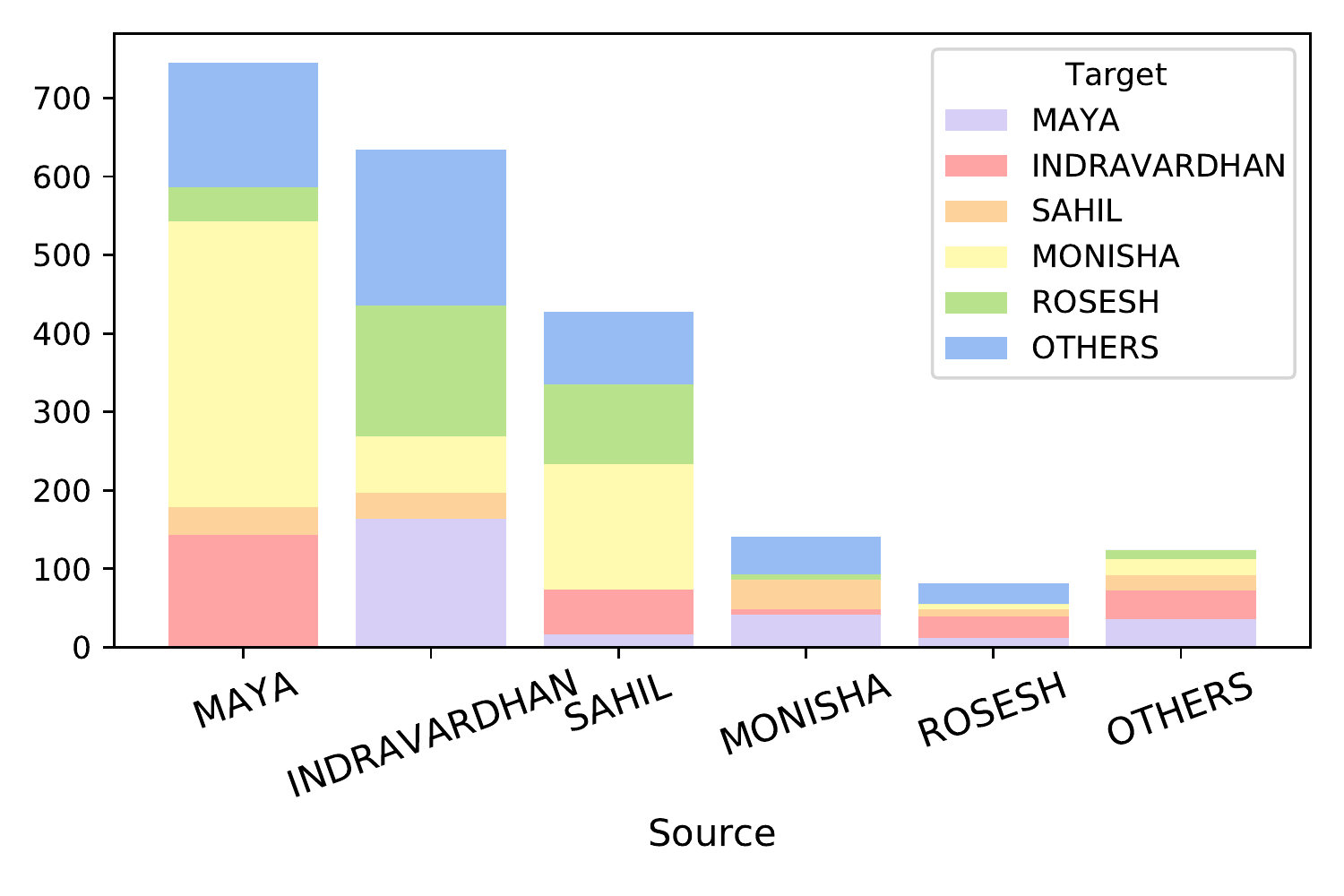}
    }}
    }\\
    \resizebox{\textwidth}{!}{
    \resizebox{0.33\textwidth}{!}{%
     \subfloat[Sarcasm source distribution]{
        \includegraphics[width=0.33\textwidth]{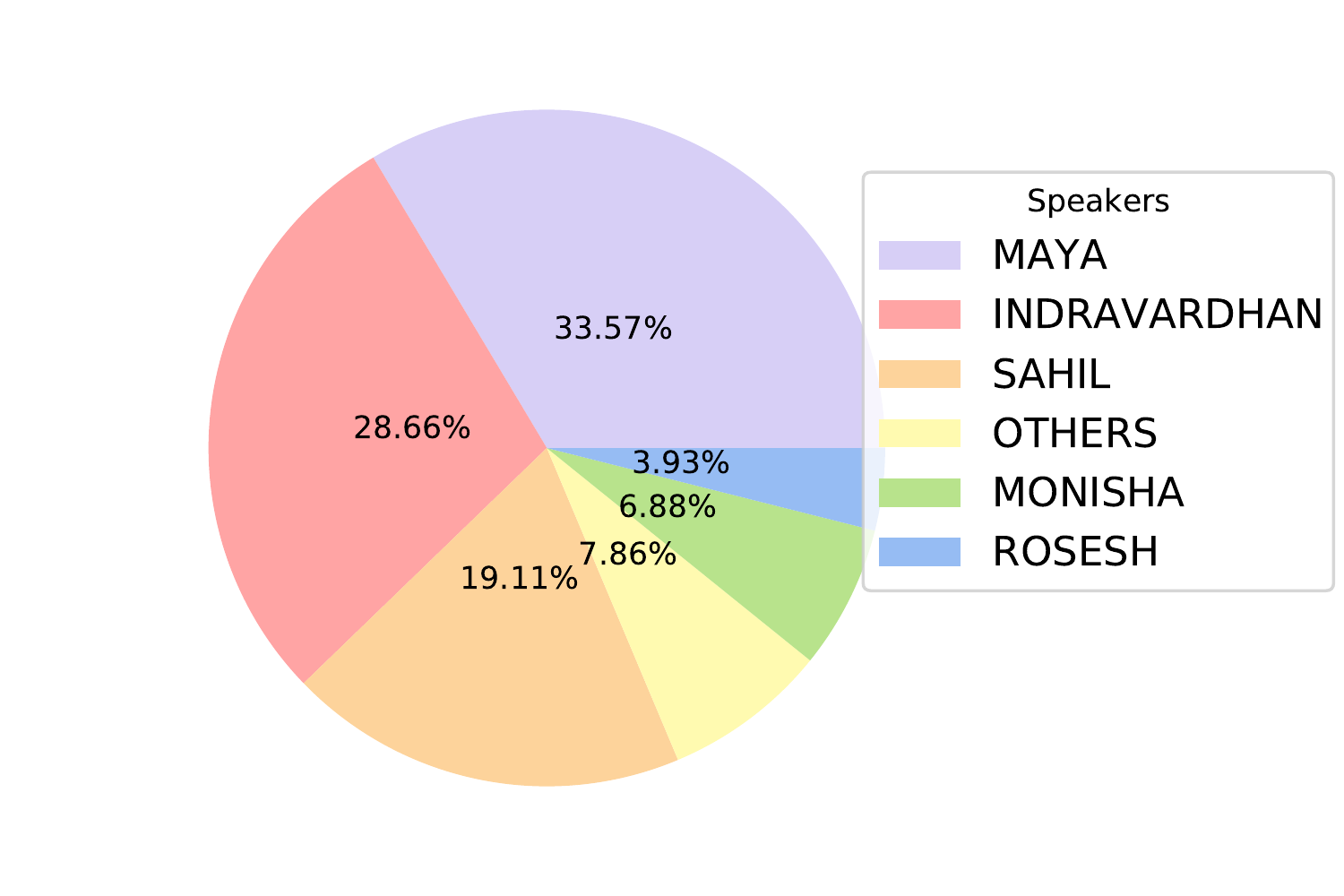}
    }}
    \resizebox{0.33\textwidth}{!}{%
    \subfloat[Sarcasm target distribution]{
        \includegraphics[width=0.33\textwidth]{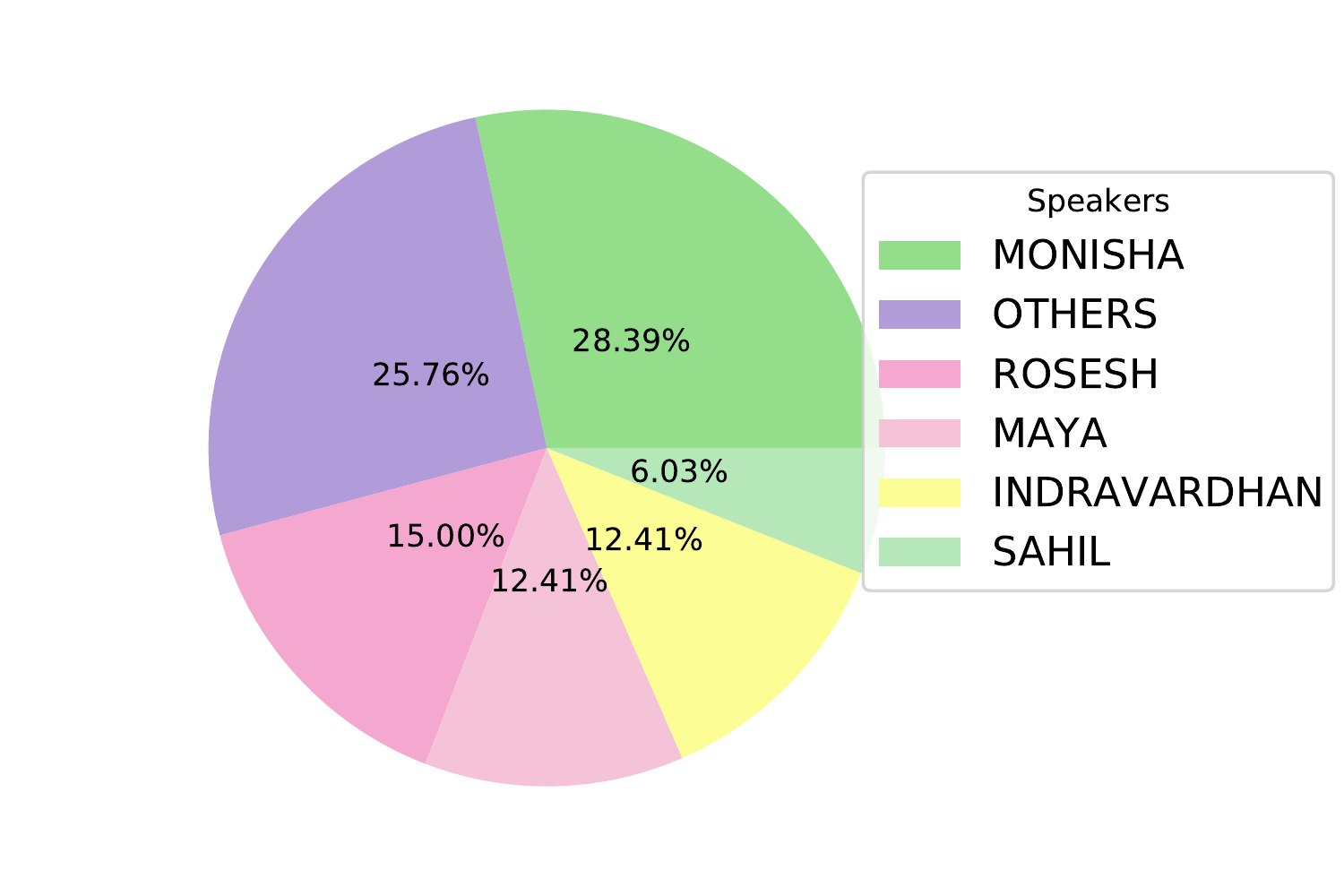}
    }}
    \resizebox{0.33\textwidth}{!}{%
    \subfloat[Explanation length distribution]{
        \includegraphics[width=0.33\textwidth]{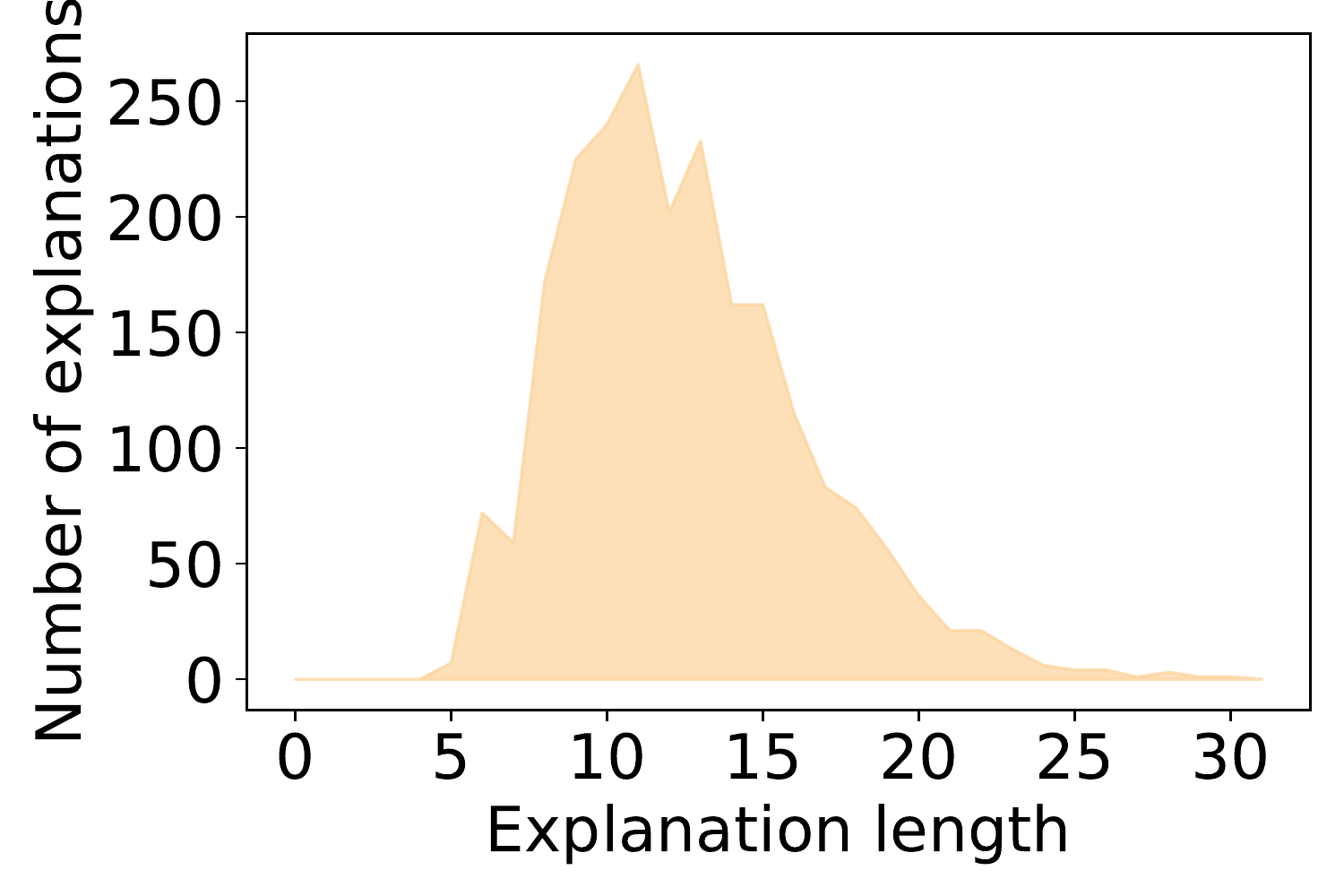}
    }}
    }
\caption{Distribution of attributes in \dataset. The number of utterances in a dialog lies between $2$ and $27$. Maximum number of speakers in a dialogue are $6$. The speaker `Maya' is the most common common sarcasm source while the speaker `Monisha' is the most prominent sarcasm target.}
\label{fig:data-stats}
\vspace{-1em}
\end{figure*}

\section{Dataset}
\label{sec:dataset}
Situational comedies, or \textit{`Sitcoms'}, vividly depict human behaviour and mannerism in everyday real-life settings. Consequently, the NLP research community has successfully used such data for sarcasm identification \citep{castro2019multimodal, 9442359}. 
However, as there is no current dataset tailored for the proposed task, we curate a new dataset named \dataset, where we augment the already existing \textsc{MaSaC} dataset \citep{9442359} with explanations for our task. \textsc{MaSaC} is a multimodal, multi-party, Hindi-English code-mixed dialogue dataset compiled from the popular Indian TV show, `Sarabhai v/s Sarabhai'\footnote{\url{https://www.imdb.com/title/tt1518542/}}. 
We manually analyze the data and clean it for our task. 
While the original dataset contained $45$ episodes of the TV series, we add $10$ more episodes along with their transcription and audio-visual boundaries. Subsequently, we select the sarcastic utterances from this augmented dataset and manually define the utterances to be included in the dialogue context for each of them. Finally, we are left with $2240$ sarcastic dialogues with the number of contextual utterances ranging from $2$ to $27$. Each of these instances is manually annotated with a corresponding natural language explanation interpreting its sarcasm. Each explanation contains four primary attributes -- source and target of sarcasm, action word for sarcasm, and an optional description for the satire as illustrated in Figure \ref{fig:se_eg}. In the explanation ``Indu implies that Maya is not looking good.", `Indu' is the sarcasm source, `Maya' is the target, `implies' is the action word, while `is not looking good' forms the description part of the explanation. We collect explanations in code-mixed format to keep consistency with the dialogue language. We split the data into train/val/test sets in an 80:10:10 ratio for our experiments, resulting in $1792$ dialogues in the train set and $224$ dialogues each in the validation and test sets. 
The next section illustrates the annotation process in more detail.
Table \ref{tab:data_stats} and Figure \ref{fig:data-stats} show detailed statistics of \dataset. 

\subsection{Annotation Guidelines}
\label{sec:app_annot}
Each of the instance in \dataset\
is associated with a corresponding video, audio, and textual transcript such that the last utterance is sarcastic in nature. We first manually define the number of contextual utterances required to understand the sarcasm present in the last utterance of each dialogue. Further, we provide each of these sarcastic statements, along with their context, to the annotators who are asked to generate an explanation for these instances based on the audio, video, and text cues. Two annotators were asked to annotate the entire dataset. The target explanation is selected by calculating the cosine similarity between the two explanations. If the cosine similarity is greater than $90\%$ then the shorter length explanation is selected as the target explanation. Otherwise, a third annotator goes through the dialogue along with the explanations and resolves the conflict. The average cosine similarity after the first pass is $87.67\%$. All the final selected explanations contain the following attributes:
\begin{itemize}[leftmargin=*,topsep=0pt]
    \setlength{\itemsep}{0pt}
    \setlength{\parskip}{0pt}
    \setlength{\parsep}{0pt}
        \item {\bf Sarcasm source:} The speaker in the dialog who is being sarcastic.
        \item {\bf Sarcasm target:} The person/ thing towards whom the sarcasm is directed.
        \item {\bf Action word:} Verb/ action used to describe how the sarcasm is taking place. For e.g. mocks, insults, taunts, etc.
        \item {\bf Description:} A description about the scene which helps in understanding the sarcasm.
\end{itemize}
Figure \ref{fig:se_eg} represents an example annotation from \dataset\ with its attributes.

\begin{figure*}[h!]
    \centering
    \includegraphics[width=\textwidth]{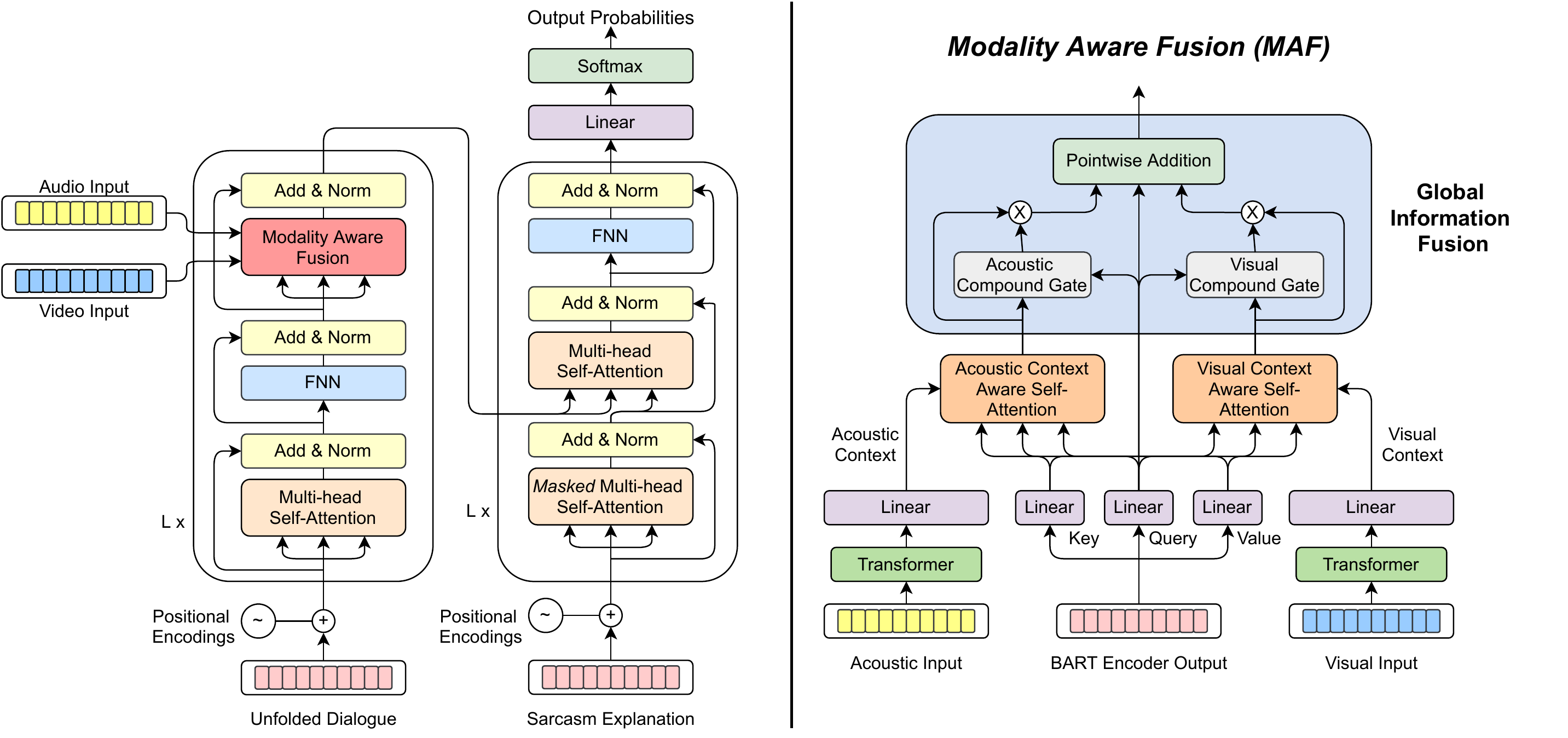}
    \caption{Model architecture for \modelTAV. The proposed Multimodal Fusion Block captures audio-visual cues using Multimodal Context Aware Attention (MCA2) which are further fused with textual representations using Global Information Fusion (GIF) block.}
    \label{fig:model_arch}
\end{figure*}

\section{Proposed Methodology}
In this section, we present our model and its nuances. The primary goal is to smoothly integrate multimodal knowledge into the BART architecture. To this end, we introduce \textit{Multimodal Aware Fusion} (\model), an adapter-based module that comprises of \textit{Multimodal Context-Aware Attention (MCA2)} and \textit{Global Information Fusion (GIF)} mechanisms. Given the textual input sarcastic dialogue along with the audio-video cues, the former aptly introduces multimodal information in the textual representations, while the latter conglomerates the audio-visual information infused textual representations. This adapter module can be readily incorporated at multiple layers of BART/mBART to facilitate various levels of multimodal interaction. Figure~\ref{fig:model_arch} illustrates our model architecture.

\subsection{Multimodal Context Aware Attention}
The traditional dot-product-based cross-modal attention scheme leads to the direct interaction of textual representations with other modalities. Here the text representations act as the query against the multimodal representations, which serve as the key and value. As each modality comes from a different embedding subspace, a direct fusion of multimodal information might not retain maximum contextual information and can also leak substantial noise in the final representations. Thus, based on the findings of \citet{Yang_Li_Wong_Chao_Wang_Tu_2019}, we propose multimodal fusion through \textit{Context Aware Attention}. We first generate multimodal information conditioned key and value vectors and then perform the traditional scaled dot-product attention. We elaborate on the process below.

Given the intermediate representation $H$ generated by the GPLMs at a specific layer, we calculate the query, key, and value vectors $Q$, $K$, and $V$ $\in \mathbb{R}^{n \times d}$, respectively, as given in Equation~\ref{eq:qkv}, where $W_Q, W_K,$ and $W_V \in \mathbb{R}^{d \times d}$ are learnable parameters. Here, $n$ denotes the maximum sequence length of the text, and $d$ denotes the dimensionality of the GPLM generated vector.
\begin{eqnarray}
    \begin{bmatrix}
    Q  K  V
    \end{bmatrix} = H \begin{bmatrix}
    W_Q  W_K  W_V
    \end{bmatrix}
    \label{eq:qkv}
\end{eqnarray}

Let $C\in \mathbb{R}^{n \times d_c}$ denote the vector obtained from audio or visual representation. We generate multimodal information informed key and value vectors $\hat K$ and $\hat V$, respectively, as given by \citet{Yang_Li_Wong_Chao_Wang_Tu_2019}. To decide how much information to integrate from the multimodal source and how much information to retain from the textual modality, we learn matrix $\lambda \in \mathbb{R}^{n \times 1}$ (Equation \ref{eq:lambda}). Note that $U_k$ and $U_v \in \mathbb{R}^{d_c \times d}$ are learnable matrices.
\begin{eqnarray}
    \begin{bmatrix}
    \hat K \\ \hat V
    \end{bmatrix} = (1 - \begin{bmatrix}
    \lambda_k \\ \lambda_v
    \end{bmatrix})\begin{bmatrix}
    K \\ V
    \end{bmatrix} + \begin{bmatrix}
    \lambda_k \\ \lambda_v
    \end{bmatrix}(C\begin{bmatrix}
    U_k \\ U_v
    \end{bmatrix})
\end{eqnarray}

Instead of making $\lambda_k$ and $\lambda_v$ as hyperparameters, we let the model decide their values using a gating mechanism as computed in Equation~\ref{eq:lambda}. The matrices of $W_{k_1}, W_{k_2}, W_{v_1},$ and $W_{v_2} \in \mathbb{R}^{d \times 1}$ are trained along with the model.

\vspace{-1em}
\begin{eqnarray}
    \begin{bmatrix}
    \lambda_k \\ \lambda_v
    \end{bmatrix} = \sigma(\begin{bmatrix}
    K \\ V
    \end{bmatrix} \begin{bmatrix}
    W_{k_1} \\ W_{v_1}
    \end{bmatrix} + C\begin{bmatrix}
    U_k \\ U_v
    \end{bmatrix} \begin{bmatrix}
    W_{k_2} \\ W_{v_2}
    \end{bmatrix})
    \label{eq:lambda}
\end{eqnarray}

Finally, the multimodal information infused vectors $\hat K$ and $\hat V$ are used to compute the traditional scaled dot-product attention. For our case, we have two modalities -- audio and video. Using the \textit{context-aware attention mechanism}, we obtain the acoustic-information-infused and visual-information infused vectors $H_A$ and $H_V$, respectively (c.f. Equations~\ref{eq:HA} and ~\ref{eq:HV}).

\vspace{-5mm}
\begin{eqnarray}
    H_a = Softmax(\frac{Q\hat K_a^T}{\sqrt{d_k}})\hat V_a 
    \label{eq:HA}\\
    H_v = Softmax(\frac{Q\hat K_v^T}{\sqrt{d_k}})\hat V_v
    \label{eq:HV}
\end{eqnarray}

\subsection{Global Information Fusion}
In order to combine the information from both the acoustic and visual modalities, we design the GIF block. We propose two gates, namely the \textit{acoustic gate}  ($g_a$) and the \textit{visual gate} ($g_v$) to control the amount of information transmitted by each modality. They are as follows:
\begin{eqnarray}
    g_a = [H \oplus H_a] W_a + b_a \\
    g_v = [H \oplus H_v] W_v + b_v
\end{eqnarray}

Here, $W_a, W_v \in \mathbb{R}^{2d \times d}$ and $b_a, b_v \in \mathbb{R}^{d \times 1}$ are trainable parameters, and $\oplus$ denotes concatenation. The final multimodal information fused representation $\hat H$ is given by Equation~\ref{eq:GIF}.
\begin{eqnarray}
\hat H = H + g_a \odot H_a + g_v \odot H_v
    \label{eq:GIF}
\end{eqnarray}

This vector $\hat H$ is inserted back into GPLM for further processing.

\section{Experiments, Results and Analysis}
In this section, we illustrate our feature extraction strategy, the comparative systems, followed by the results and its analysis. For a quantitative analysis of the generated explanations,  we use the standard metrics for generative tasks -- ROUGE-1/2/L \cite{lin-2004-rouge}, BLEU-1/2/3/4 \cite{papineni2002bleu}, and METEOR \cite{denkowski:lavie:meteor-wmt:2014}. To capture the semantic similarity, we use the multilingual version of the BERTScore \citep{zhang2019bertscore}.

\subsection{Feature Extraction}
    \paragraph{Audio:} Acoustic representations for each instance are obtained using the openSMILE python library\footnote{\url{https://audeering.github.io/opensmile-python/}}. We use a window size of $25$ ms and a window shift of $10$ ms to get the non-overlapping frames. Further, we employ the eGeMAPS model \cite{eyben2015geneva} and extract $154$ dimensional functional features such as Mel Frequency Cepstral Coefficients (MFCCs) and loudness for each frame of the instance. These features are then fed to a Transformer encoder \cite{vaswani2017attention} for further processing.
    \paragraph{Video:} We use a pre-trained action recognition model, ResNext-101 \cite{hara2018spatiotemporal}, trained on the Kinetics dataset \cite{kay2017kinetics} which can recognise $101$ different actions. We use a frame rate of $1.5$, a resolution of $720$ pixels, and a window length of $16$  to extract the $2048$ dimensional visual features. Similar to audio feature extraction, we employ a Transformer encoder \cite{vaswani2017attention} to capture the sequential dialogue context in the representations.
    
    \subsection{Comparative Systems}
        To get the best textual representations for the dialogues, we experiment with various sequence-to-sequence (seq2seq) architectures. \textbf{RNN:} We use the openNMT\footnote{\url{https://github.com/OpenNMT/OpenNMT-py}} implementation of the RNN seq-to-seq architecture.
        \textbf{Transformer} \cite{vaswani2017attention}: The standard Transformer encoder and decoder are used to generate explanations in this case. 
        \textbf{Pointer Generator Network} \cite{see2017get}: A seq-to-seq architecture that allows the generation of new words as well as copying words from the input text for generating accurate summaries. \textbf{BART} \cite{lewis2019bart}: It is a denoising auto-encoder model with standard machine translation architecture with a bidirectional encoder and an auto-regressive left-to-right decoder. We use its base version.
        \textbf{mBART} \cite{liu2020multilingual}: Following the same architecture and objective as BART, mBART is trained on large-scale monolingual corpora in different languages \footnote{\url{https://huggingface.co/facebook/mbart-large-50-many-to-many-mmt}}.
 
\subsection{Results}
\paragraph{Text Based:} As evident from Table~\ref{tab:all_results}, BART performs the best across all the metrics for the textual modality, showing an improvement of almost $2$-$3\%$ on the METEOR and ROUGE scores when compared with the next best baseline. PGN, RNN, and Transformers demonstrate admissible performance considering that they have been trained from scratch. However, it is surprising to see mBART not performing better than BART as it is trained on multilingual data.  We elaborate more on this in Appendix ~\ref{sec:bart_vs_mbart}. 

\begin{table}[]
\centering
\resizebox{\columnwidth}{!}{%
\begin{tabular}{|l|l|c|c|c|c|c|c|c|c|c|}
\hline
\textbf{Mode} & \textbf{Model} & \textbf{R1} & \textbf{R2} & \textbf{RL} & \textbf{B1} & \textbf{B2} & \textbf{B3} & \textbf{B4} & \textbf{M} & \textbf{BS} \\ \hline \hline
\multirow{5}{*}{\rotatebox{90}{\textbf{Textual}}} & \textbf{RNN} & 29.22 & 7.85 & 27.59 & 22.06 & 8.22 & 4.76 & 2.88 & 18.45 & 73.24 \\ 
 & \textbf{Transformers} & 29.17 & 6.35 & 27.97 & 17.79 & 5.63 & 2.61 & 0.88 & 15.65 & 72.21 \\ 
 & \textbf{PGN} & 23.37 & 4.83 & 17.46 & 17.32 & 6.68 & 1.58 & 0.52 & 23.54 & 71.90 \\ 
 & \textbf{mBART} & 33.66 & 11.02 & 31.50 & 22.92 & 10.56 & 6.07 & 3.39 & 21.03 & 73.83 \\ 
 & \textbf{BART} & 36.88 & 11.91 & 33.49 & 27.44 & 12.23 & 5.96 & 2.89 & 26.65 & 76.03 \\ \hline
\multirow{6}{*}{\rotatebox{90}{\textbf{Multimodality}}} & \textbf{\modelTAm} & 39.02 & 15.90 & 36.83 & 31.26 & 16.94 & 11.54 & 7.72 & 29.05 & 77.06 \\ 
& \textbf{\modelTVm} & 39.47 & 16.78 & \bf 37.38 & 32.44 & 17.91 & 12.02 & 7.36 & 29.74 & 77.47 \\ 
& \cellcolor{LightCyan} \textbf{\modelTAVm} & \cellcolor{LightCyan} 38.52 & \cellcolor{LightCyan} 14.13 & \cellcolor{LightCyan} 36.60 & \cellcolor{LightCyan} 30.50 & \cellcolor{LightCyan} 15.20 & \cellcolor{LightCyan} 9.78 & \cellcolor{LightCyan} 5.74 & \cellcolor{LightCyan} 27.42 & \cellcolor{LightCyan} 76.70 \\ \cline{2-11} 
& \textbf{\modelTA} & 38.21 & 14.53 & 35.97 & 30.58 & 15.36 & 9.63 & 5.96 & 27.71 & 77.08 \\ 
& \textbf{\modelTV} & 37.48 & 15.38 & 35.64 & 30.28 & 16.89 & 10.33 & 6.55 & 28.24 & 76.95  \\ 
& \cellcolor{LightCyan} \textbf{\modelTAV} & \cellcolor{LightCyan} \bf 39.69 & \cellcolor{LightCyan} \bf 17.10 & \cellcolor{LightCyan} 37.37 & \cellcolor{LightCyan} \bf 33.20 & \cellcolor{LightCyan} \bf 18.69 & \cellcolor{LightCyan} \bf 12.37 & \cellcolor{LightCyan} \bf 8.58 & \cellcolor{LightCyan} \bf 30.40 & \cellcolor{LightCyan} \bf 77.67 \\ \hline
\end{tabular}%
}
\caption{Experimental results. (Abbreviation: R1/2/L: ROUGE1/2/L; B1/2/3/4: BLEU1/2/3/4; M: METEOR; BS: BERT Score; PGN: Pointer Generator Network).}
\label{tab:all_results}
\vspace{-4mm}
\end{table}

\paragraph{Multimodality:} Psychological and linguistic literature suggests that there exist distinct paralinguistic cues that aid in comprehending sarcasm and humour \citep{attardo2003multimodal, Tabacaru_Lemmens_2014}. Thus, we gradually merge auditory and visual modalities using \model\ module and obtain \modelTAV\ and \modelTAVm\ for BART and mBART, respectively.
We observe that the inclusion of acoustic signals leads to noticeable gains of $2$-$3\%$ across the ROUGE, BLEU, and METEOR scores. The rise in BERTScore also suggests that the multimodal variant generates a tad more coherent explanations. As ironical intonations such as mimicry, monotone, flat contour, extremes of pitch, long pauses, and exaggerated pitch \citep{rockwell2007vocal} form a significant component in sarcasm understanding, we surmise that our model, to some extent, is able to spot such markers and identify the intended sarcasm behind them. 

We notice that visual information also contributes to our cause. Significant performance gains are observed for \modelTV\ and \modelTVm, as all the metrics show a rise of about $3$-$4\%$. While \modelTA\ gives marginally better performance over \modelTV\ in terms of R1, RL, and B1, we see that \modelTV\ performs better in terms of the rest of the metrics. Often, sarcasm is depicted through gestural cues such as raised eyebrows, a straight face, or an eye roll \citep{attardo2003multimodal}. Moreover, when satire is conveyed by mocking someone's looks or physical appearances, it becomes essential to incorporate information expressed through visual media. Thus, we can say that, to some extent, our model is able to capture these nuances of non-verbal cues and use them well to normalize the sarcasm in a dialogue. In summary, we conjecture that whether independent or together, audio-visual signals bring essential information to the table for understanding sarcasm.

\subsection{Ablation Study}
\begin{table}[]
\centering
\resizebox{\columnwidth}{!}{%
\begin{tabular}{|l|c|c|c|c|c|c|c|c|c|}
\hline
\bf Model & \bf R1 & \bf R2 & \bf RL & \bf B1 & \bf B2 & \bf B3 & \bf B4 & \bf M & \bf BS \\ \hline \hline

\rowcolor{LightCyan} \bf \modelTAVm & \bf 38.52 & 14.13 & 36.60 & 30.50 & 15.20 & 9.78 & 5.74 & 27.42 & 76.70 \\

\bf \quad - MCA2 + \textsc{Concat1} & 37.56  & \bf 14.85 & 34.90 & 30.16  & 15.76  & \bf 10.12  & \bf 6.82 & \bf 28.59 & 76.59\\

\bf \quad - MAF + \textsc{Concat2} & 17.22  & 1.70  & 14.12  & 13.11  & 2.11  & 0.00  & 0.00  & 9.34 & 66.64\\

\bf \quad - MCA2 + \textsc{DPA} & 36.43 & 13.04  & 33.75  & 28.73  & 14.02  & 8.00 & 4.89 & 25.60 & 75.58 \\

\bf \quad - GIF & 36.37  & 13.85 & 34.92 & 28.49 & 14.34 & 9.00  & 6.16 & 25.75 & \bf 76.86 \\ \hline

\rowcolor{LightCyan} \bf \modelTAV & \bf 39.69 & \bf 17.10 & \bf 37.37 & \bf 33.20 & \bf 18.69 & \bf 12.37 & \bf 8.58 & \bf 30.40 & \bf 77.67 \\

\bf \quad - MCA2 + \textsc{Concat1} & 36.88 & 13.21 & 34.39 & 29.63 & 14.56 & 8.43 & 4.84 & 26.15 & 76.08 \\

\bf \quad - MAF + \textsc{Concat2} & 21.11 & 2.31 & 19.68 & 12.44 & 2.44 & 0.73 & 0.31 & 9.51 & 69.54 \\

\bf \quad - MCA2 + \textsc{DPA} & 38.84 & 14.76 & 36.96 & 30.23 & 15.95 & 9.88 & 5.83 & 28.04 & 77.20 \\

\bf \quad - GIF & 39.45 & 14.85 & 37.18 & 31.85 & 15.97 & 9.62 & 5.47 & 28.87 & 77.54\\ \hline
\end{tabular}%
}
\caption{Ablation results on \modelTAVm\ and \modelTAV\ (DPA: Dot Product Attention).}
\label{tab:ablation}
\vspace{-3mm}
\end{table}

Table \ref{tab:ablation} reports the ablation study. \textsc{Concat1} represents the case where we perform bimodal concatenation ($(T\oplus A), (T\oplus V)$) instead of the \textit{MCA2} mechanism, followed by the GIF module, whereas, \textsc{Concat2} represents the simple trimodal concatenation ($T\oplus A \oplus V$) of acoustic, visual, and textual representations followed by a linear layer for dimensionality reduction. In comparison with \textit{MCA2}, \textsc{Concat2} reports a below-average performance with a significant drop of more than $14\%$ for \modelTAV\ and \modelTAVm. This highlights the need to have deftly crafted multimodal fusion mechanisms. \textsc{Concat1}, on the other hand, gives good performance and is competitive with \textsc{DPA} and \modelTAV. We speculate that treating the audio and video modalities separately and then merging them to retain the complimentary and differential features lead to this performance gain. Our proposed \model\ outperforms \textsc{DPA} with gains of $1$-$3\%$. This underlines that our unique multimodal fusion strategy is aptly able to capture the contextual information provided by the audio and video signals. Replacing the \textit{GIF} module with simple addition, we observe a noticeable decline in the performance across almost all metrics by about $2$-$3\%$. This attests to the inclusion of \textit{GIF} module over simple addition.
We also experiment with fusing multimodal information using \model\ before different layers of the BART encoder. The best performance was obtained when the fusion was done before the sixth layer of the architecture (c.f. Appendix \ref{sec:app_fusion}).

\begin{table*}[ht]
\centering
\subfloat[Incoherent explanation]{
\label{tab:coherence}
\resizebox{0.33\textwidth}{!}{%
\begin{tabular}{p{3em}p{20em}}
\multicolumn{2}{c}{\includegraphics{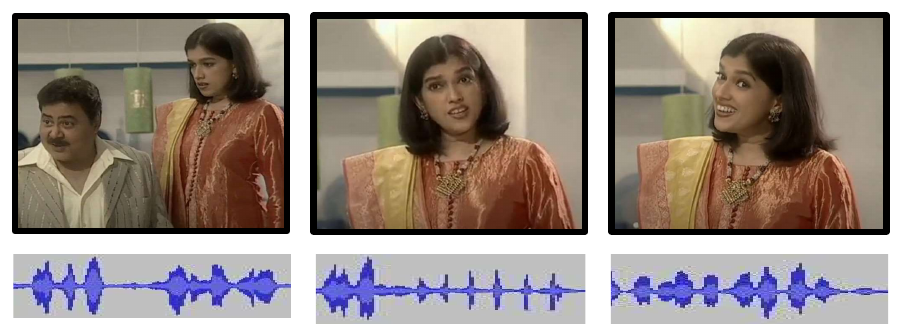}} \\ \hline
\multicolumn{2}{l}{\begin{tabular}[c]{@{}p{25em}@{}}{INDRAVARDHAN:} Accha suno Monisha tumhaare ghar mein been ya aisa kuuch hain? \textit{\color{blue}Listen Monisha, do you have a flute or something similar?}\\{MAYA:} Kaise hogi? Monisha aapne ghar pe dustbin mushkil se rakhti hain to snake charmer waali been kaha se rakhegi? \textit{\color{blue}How will it be there? Monisha hardly keeps a dustbin in her home so how will she has a snake charmer's flute?}\end{tabular}} \\ \hline
\rowcolor{LightGreen}\bf Gold & Maya Monisha ko tana marti hai safai ka dhyan na rakhne ke liye \textit{\color{blue}Maya taunts Monisha for not keeping a check of cleanliness}\\ 
\bf BART & Maya Monisha ko tumaari burayi nahi karta. \textit{\color{blue} Maya doesn't blame you for Monisha}\\
& \\
& \\
\bf \modelTAV & Maya implies ki Monisha bohot ghar mein bahar nahi kar sakati. \textit{\color{blue}Maya implies that Monisha very in home cannot do outside.}\\
\hline
\end{tabular}
}
}
\subfloat[Explanation related to dialogue]{
\label{tab:context}
\resizebox{0.33\textwidth}{!}{%
\begin{tabular}{p{3em}p{20em}} 
\multicolumn{2}{c}{\includegraphics{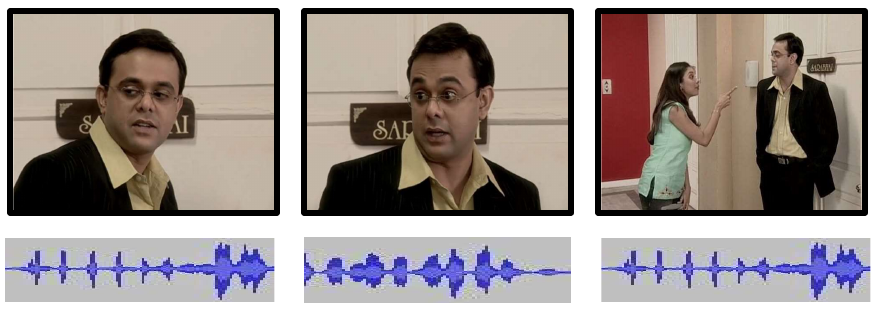}} \\ \hline
\multicolumn{2}{l}{\begin{tabular}[c]{@{}p{25em}@{}}SAHIL: Ab tumne ghar ki itni saaf safai ki hai and secondly us Karan Verma ke liye pasta, lasagne, caramel custard banaya. \textit{\color{blue}Now you have cleaned the house so much and secondly made pasta, lasagne, caramel custard for that Karan Verma.}\\MONISHA: Walnut brownie bhi. \textit{\color{blue}And walnut brownie too.}\\SAHIL: Walnut brownie, matlab wo khane wali? \textit{\color{blue}You mean edible walnut brownie?}\end{tabular}} \\
\hline
\rowcolor{LightGreen} \bf Gold & Sahil monisha ki cooking ka mazak udata hai \textit{\color{blue}Sahil makes fun of Monisha's cooking.}\\ 
\rowcolor{LightGreen} & \\
\bf BART & Monisha sahil ko walnut brownie ki matlab wo khane wali. \textit{\color{blue}Walnut Brownie to Monisha Sahil means she eats}\\
& \\
\bf \modelTAV & Sahil monisha ki cooking ka mazak udata hai \textit{\color{blue}Sahil makes fun of Monisha's cooking.}\\
& \\
\hline
\end{tabular}
}
}
\subfloat[Explanation related to sarcasm]{
\label{tab:sarcasm}
\resizebox{0.33\textwidth}{!}{%
\begin{tabular}{p{3em}p{20em}}
\multicolumn{2}{c}{\includegraphics{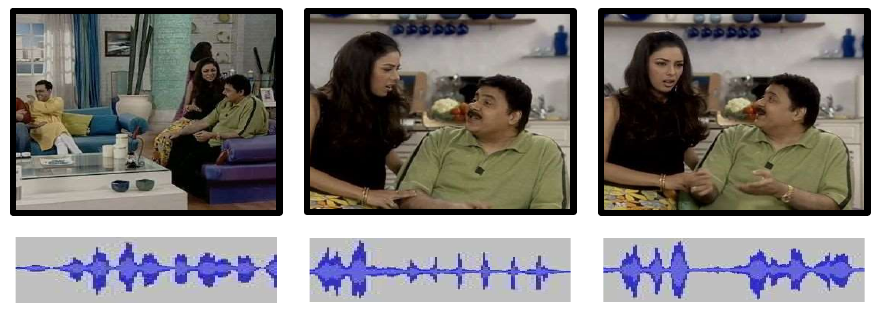}} \\ \hline
\multicolumn{2}{l}{\begin{tabular}[c]{@{}p{25em}@{}}MONISHA: Ladki ka naam Ajanta Kyon Rakha? \textit{\color{blue}Why did they named the girl Ajanta?}\\INDRAVARDHAN: Kyunki uski maa ajanta caves dekh rahi thi Jab vo Paida Hui haha. \textit{\color{blue}Because her mother must be watching the Ajanta caves when she was born haha.}\end{tabular}}\\
& \\
& \\
\hline
\rowcolor{LightGreen} \bf Gold & Indravadan Ajanta ke naam ka mazak udata hai \textit{\color{blue}Indravardhan makes fun of Ajanta's name} \\
\rowcolor{LightGreen} & \\
\bf BART & Indravardhan Monisha ko taunt maarta hai ki uski maa ajanta caves dekh rahi thi Jab vo Paida Hui \textit{\color{blue}
Indravardhan taunts Monisha as her mother was watching Ajanta Caves when she was born.}\\
\bf \modelTAV & Indravadan ajanta ke naam ka mazak udata hai \textit{\color{blue}Indravardhan makes fun of Ajanta's name}\\
& \\
\hline
\end{tabular}
}
}
\caption{Actual and generated explanations for sample dialogues from test set. The last utterance is the sarcastic utterance for each dialogue.}
\label{tab:qual_analysis}
\end{table*}
\subsection{Result Analysis}
We evaluate the generated explanations based on their ability to correctly identify the source and target of a sarcastic comment in a conversation. 
We report such results for mBART, BART, \modelTA, \modelTV, and \modelTAV. BART performs better than mBART for the source as well as target identification. We observe that the inclusion of audio $(\uparrow10\%)$ and video $(\uparrow8\%)$ information drastically improves the source identification capability of the model. The combination of both these non-verbal cues leads to a whopping improvement of more than $13\%$ for the same. As a result, we infer that multimodal fusion enables the model to incorporate audio-visual peculiarities unique to each speaker, resulting in improved source identification. The performance for target identification, however, drops slightly on the inclusion of multimodality. We encourage future work in this direction.

\begin{table}[]
\centering
\resizebox{\columnwidth}{!}{%
\begin{tabular}{|l|c|c|c|c|c|}
\hline
 & \textbf{mBART} & \textbf{BART} & \textbf{\modelTA} & \textbf{\modelTV} & \textbf{\modelTAV} \\ \hline \hline
\textbf{Source} & 75.00 & 77.23 & 87.94 & 85.71 & \bf 91.07 \\
\textbf{Target} & 45.53 & \bf 52.67 & 43.75 & 43.75 & 46.42 \\ \hline
\end{tabular}%
}
\caption{Source-target accuracy of the generated explanations for BART-based systems.}
\label{tab:analysis}
\vspace{-3mm}
\end{table}

\paragraph{Qualitative Analysis.}
We analyze the best performing model, \modelTAV, and its corresponding unimodal model, BART, and present some examples in Table \ref{tab:qual_analysis}. In Table \ref{tab:coherence}, we show one instance where the explanations generated by the BART as well as \modelTAV\ are neither coherent nor comply with the dialogue context and contain much scope of improvement. On the other hand, Table \ref{tab:context} illustrates an instance where the explanation generated by \modelTAV\ adheres to the topic of the dialogue, unlike the one generated by its unimodal counterpart. Table \ref{tab:sarcasm} depicts a dialogue where \modelTAV's explanation better captures the satire than BART. 
We further dissect the models based on different modalities in Appendix \ref{sec:app_qual_analysis}. 
\paragraph{Human Evaluation.}
Since the proposed \task\ task is a generative task, it is imperative to manually inspect the generated results. Consequently, we perform a human evaluation for a sample of $30$ instances from our test set with the help of $25$ evaluators\footnote{Evaluators are the experts in linguistics and NLP and their age ranges in 20-28 years.}. We ask the evaluators to judge the generated explanation, given the transcripts of the sarcastic dialogues along with a small video clip with audio as well. Each evaluator has to see the video clips and then rate the generated explanations on a scale of $0$ to $5$ based on the following factors\footnote{$0$ denoting poor performance while $5$ signifies perfect performance.}:
\begin{itemize}[leftmargin=*,topsep=0pt]
\setlength{\itemsep}{0pt}
\setlength{\parskip}{0pt}
\setlength{\parsep}{0pt}
    \item \textbf{Coherence:} Measures how well the explanations are organized and structured.
    \item \textbf{Related to dialogue:} Measures whether the generated explanation adheres to the topic of the dialogue.
    \item \textbf{Related to sarcasm:} Measures whether the explanation is talking about something related to the sarcasm present in the dialogue.
\end{itemize}
Table \ref{tab:human_eval} presents the human evaluation analysis with average scores for each of the aforementioned categories. Our scrutiny suggests that \modelTAV\ generates more syntactically coherent explanations when compared with its textual and bimodal counterparts. Also, \modelTAV\ and \modelTV\ generate explanations that are more focused on the conversation's topic, as we see an increase of $0.55$ points in the \textit{related to the dialogue} category. Thus, we reestablish that these models are able to incorporate information that is explicitly absent from the dialogue, such as scene description, facial features, and looks of the characters. Furthermore, we establish that \modelTAV\ is better able to grasp sarcasm and its normalization, as it shows about $0.6$ points improvement over BART in the \textit{related to sarcasm} category. Lastly, as none of the metrics in Table~\ref{tab:human_eval} exhibit high scores ($3.5+$), we feel there is still much scope for improvement in terms of the generation performance and human evaluation. The research community can further explore the task with our proposed dataset, \dataset.
\begin{table}[t!]
\centering
\resizebox{\columnwidth}{!}{%
\begin{tabular}{|l|c|c|c|}
\hline
 & \textbf{Coherency} & \textbf{Related to dialogue} & \textbf{Related to sarcasm} \\ \hline \hline
\textbf{mBART} & 2.57 & 2.66 & 2.15 \\ \hline
\textbf{BART} & 2.73 & 2.56 & 2.18 \\ \hline
\textbf{\modelTA} & 2.95 & 2.91 & 2.51 \\ \hline
\textbf{\modelTV} & 3.01 & \bf 3.11 & 2.66 \\ \hline
\textbf{\modelTAV} & \bf 3.03 & \bf 3.11 & \bf 2.77 \\ \hline
\end{tabular}%
}
\caption{Human evaluation statistics -- comparing different models. Multimodal models are BART based.}
\label{tab:human_eval}
\vspace{-5mm}
\end{table}

\section{Conclusion}
In this work, we proposed the new task of \textit{\textbf{Sarcasm Explanation in Dialogue ({\task}})}, which aims to generate a natural language explanation for sarcastic conversations. We curated {\dataset}, a novel multimodal, multiparty, code-mixed, dialogue dataset to support the \task\ task. We experimented with multiple text and multimodal baselines, which give promising results on the task at hand. Furthermore, we designed a unique multimodal fusion scheme to merge the textual, acoustic, and visual features via  Multimodal Context-Aware Attention (MCA2) and  Global Information Fusion (GIF) mechanisms. As hypothesized, the results show that acoustic and visual features support our task and thus, generate better explanations. We show extensive qualitative analysis of the explanations obtained from different models and highlight their advantages as well as pitfalls. We also perform a thorough human evaluation to compare the performance of the models with that of human understanding. Though the models augmented with the proposed fusion strategy perform better than the rest, the human evaluation suggested there is still room for improvement which can be further explored in future studies.


\section*{Acknowledgement}
The authors would like to acknowledge the support of the Ramanujan Fellowship (SERB, India), Infosys Centre for AI (CAI) at IIIT-Delhi, and ihub-Anubhuti-iiitd Foundation set up under the NM-ICPS scheme of the Department of Science and Technology, India.

\bibliography{anthology,custom}

\begin{thebibliography}{49}
\expandafter\ifx\csname natexlab\endcsname\relax\def\natexlab#1{#1}\fi

\bibitem[{Abu~Farha and Magdy(2020)}]{abu-farha-magdy-2020-arabic}
Ibrahim Abu~Farha and Walid Magdy. 2020.
\newblock \href {https://aclanthology.org/2020.osact-1.5} {From {A}rabic
  sentiment analysis to sarcasm detection: The {A}r{S}arcasm dataset}.
\newblock In \emph{Proceedings of the 4th Workshop on Open-Source Arabic
  Corpora and Processing Tools, with a Shared Task on Offensive Language
  Detection}, pages 32--39, Marseille, France. European Language Resource
  Association.

\bibitem[{Attardo et~al.(2003)Attardo, Eisterhold, Hay, and
  Poggi}]{attardo2003multimodal}
Salvatore Attardo, Jodi Eisterhold, Jennifery Hay, and Isabella Poggi. 2003.
\newblock \href {https://psycnet.apa.org/doi/10.1515/humr.2003.012} {Multimodal
  markers of irony and sarcasm.}
\newblock \emph{Humor: International Journal of Humor Research}, 16(2).

\bibitem[{Bedi et~al.(2021)Bedi, Kumar, Akhtar, and Chakraborty}]{9442359}
Manjot Bedi, Shivani Kumar, Md~Shad Akhtar, and Tanmoy Chakraborty. 2021.
\newblock \href {https://doi.org/10.1109/TAFFC.2021.3083522} {Multi-modal
  sarcasm detection and humor classification in code-mixed conversations}.
\newblock \emph{IEEE Transactions on Affective Computing}, pages 1--1.

\bibitem[{Bharti et~al.(2017)Bharti, Sathya~Babu, and
  Jena}]{bharti2017harnessing}
Santosh~Kumar Bharti, Korra Sathya~Babu, and Sanjay~Kumar Jena. 2017.
\newblock \href {https://doi.org/https://doi.org/10.1007/978-3-319-69900-4_86}
  {Harnessing online news for sarcasm detection in hindi tweets}.
\newblock In \emph{Pattern Recognition and Machine Intelligence}, pages
  679--686, Cham. Springer International Publishing.

\bibitem[{Cai et~al.(2019)Cai, Cai, and Wan}]{cai-etal-2019-multi}
Yitao Cai, Huiyu Cai, and Xiaojun Wan. 2019.
\newblock \href {https://doi.org/10.18653/v1/P19-1239} {Multi-modal sarcasm
  detection in {T}witter with hierarchical fusion model}.
\newblock In \emph{Proceedings of the 57th Annual Meeting of the Association
  for Computational Linguistics}, pages 2506--2515, Florence, Italy.
  Association for Computational Linguistics.

\bibitem[{Castro et~al.(2019)Castro, Hazarika, P{\'e}rez-Rosas, Zimmermann,
  Mihalcea, and Poria}]{castro2019multimodal}
Santiago Castro, Devamanyu Hazarika, Ver{\'o}nica P{\'e}rez-Rosas, Roger
  Zimmermann, Rada Mihalcea, and Soujanya Poria. 2019.
\newblock \href {https://doi.org/10.18653/v1/P19-1455} {Towards multimodal
  sarcasm detection (an {\_}{O}bviously{\_} perfect paper)}.
\newblock In \emph{Proceedings of the 57th Annual Meeting of the Association
  for Computational Linguistics}, pages 4619--4629, Florence, Italy.
  Association for Computational Linguistics.

\bibitem[{Chakrabarty et~al.(2020)Chakrabarty, Ghosh, Muresan, and
  Peng}]{chakrabarty-etal-2020-r}
Tuhin Chakrabarty, Debanjan Ghosh, Smaranda Muresan, and Nanyun Peng. 2020.
\newblock \href {https://doi.org/10.18653/v1/2020.acl-main.711} {{R}{\^{}}3:
  Reverse, retrieve, and rank for sarcasm generation with commonsense
  knowledge}.
\newblock In \emph{Proceedings of the 58th Annual Meeting of the Association
  for Computational Linguistics}, pages 7976--7986, Online. Association for
  Computational Linguistics.

\bibitem[{Chauhan et~al.(2020)Chauhan, S~R, Ekbal, and
  Bhattacharyya}]{chauhan-etal-2020-sentiment}
Dushyant~Singh Chauhan, Dhanush S~R, Asif Ekbal, and Pushpak Bhattacharyya.
  2020.
\newblock \href {https://doi.org/10.18653/v1/2020.acl-main.401} {Sentiment and
  emotion help sarcasm? a multi-task learning framework for multi-modal
  sarcasm, sentiment and emotion analysis}.
\newblock In \emph{Proceedings of the 58th Annual Meeting of the Association
  for Computational Linguistics}, pages 4351--4360, Online. Association for
  Computational Linguistics.

\bibitem[{Cignarella et~al.(2018)Cignarella, Frenda, Basile, Bosco, Patti,
  Rosso et~al.}]{cignarella2018overview}
Alessandra~Teresa Cignarella, Simona Frenda, Valerio Basile, Cristina Bosco,
  Viviana Patti, Paolo Rosso, et~al. 2018.
\newblock \href {http://ceur-ws.org/Vol-2263/paper005.pdf} {Overview of the
  evalita 2018 task on irony detection in italian tweets (ironita)}.
\newblock In \emph{Sixth Evaluation Campaign of Natural Language Processing and
  Speech Tools for Italian (EVALITA 2018)}, volume 2263, pages 1--6. CEUR-WS.

\bibitem[{Colston(1997)}]{colston1997salting}
Herbert~L. Colston. 1997.
\newblock \href {https://doi.org/10.1080/01638539709544980} {Salting a wound or
  sugaring a pill: The pragmatic functions of ironic criticism}.
\newblock \emph{Discourse Processes}, 23(1):25--45.

\bibitem[{Colston and Keller(1998)}]{colston1998you}
Herbert~L Colston and Shauna~B Keller. 1998.
\newblock \href {https://doi.org/https://doi.org/10.1023/A:1023229304509}
  {You'll never believe this: Irony and hyperbole in expressing surprise}.
\newblock \emph{Journal of psycholinguistic research}, 27(4):499--513.

\bibitem[{Denkowski and Lavie(2014)}]{denkowski:lavie:meteor-wmt:2014}
Michael Denkowski and Alon Lavie. 2014.
\newblock \href {https://doi.org/10.3115/v1/W14-3348} {Meteor universal:
  Language specific translation evaluation for any target language}.
\newblock In \emph{Proceedings of the Ninth Workshop on Statistical Machine
  Translation}, pages 376--380, Baltimore, Maryland, USA. Association for
  Computational Linguistics.

\bibitem[{Dubey et~al.(2019)Dubey, Joshi, and Bhattacharyya}]{dubey2019deep}
Abhijeet Dubey, Aditya Joshi, and Pushpak Bhattacharyya. 2019.
\newblock \href {https://doi.org/10.1145/3297001.3297043} {Deep models for
  converting sarcastic utterances into their non sarcastic interpretation}.
\newblock In \emph{Proceedings of the ACM India Joint International Conference
  on Data Science and Management of Data}, CoDS-COMAD '19, page 289–292, New
  York, NY, USA. Association for Computing Machinery.

\bibitem[{Eyben et~al.(2016)Eyben, Scherer, Schuller, Sundberg, André, Busso,
  Devillers, Epps, Laukka, Narayanan, and Truong}]{eyben2015geneva}
Florian Eyben, Klaus~R. Scherer, Björn~W. Schuller, Johan Sundberg, Elisabeth
  André, Carlos Busso, Laurence~Y. Devillers, Julien Epps, Petri Laukka,
  Shrikanth~S. Narayanan, and Khiet~P. Truong. 2016.
\newblock \href {https://doi.org/10.1109/TAFFC.2015.2457417} {The geneva
  minimalistic acoustic parameter set (gemaps) for voice research and affective
  computing}.
\newblock \emph{IEEE Transactions on Affective Computing}, 7(2):190--202.

\bibitem[{Ghosh et~al.(2018)Ghosh, Fabbri, and Muresan}]{10.1162/coli_a_00336}
Debanjan Ghosh, Alexander~R. Fabbri, and Smaranda Muresan. 2018.
\newblock \href {https://doi.org/10.1162/coli_a_00336} {Sarcasm analysis using
  conversation context}.
\newblock \emph{Computational Linguistics}, 44(4):755--792.

\bibitem[{Ghosh et~al.(2017)Ghosh, Richard~Fabbri, and Muresan}]{ghosh2017role}
Debanjan Ghosh, Alexander Richard~Fabbri, and Smaranda Muresan. 2017.
\newblock \href {https://doi.org/10.18653/v1/W17-5523} {The role of
  conversation context for sarcasm detection in online interactions}.
\newblock In \emph{Proceedings of the 18th Annual {SIG}dial Meeting on
  Discourse and Dialogue}, pages 186--196, Saarbr{\"u}cken, Germany.
  Association for Computational Linguistics.

\bibitem[{Hara et~al.(2018)Hara, Kataoka, and Satoh}]{hara2018spatiotemporal}
Kensho Hara, Hirokatsu Kataoka, and Yutaka Satoh. 2018.
\newblock \href
  {https://openaccess.thecvf.com/content_cvpr_2018/html/Hara_Can_Spatiotemporal_3D_CVPR_2018_paper.html}
  {Can spatiotemporal 3d cnns retrace the history of 2d cnns and imagenet?}
\newblock In \emph{Proceedings of the IEEE Conference on Computer Vision and
  Pattern Recognition (CVPR)}.

\bibitem[{Hasan et~al.(2021)Hasan, Lee, Rahman, Zadeh, Mihalcea, Morency, and
  Hoque}]{Hasan_Lee_Rahman_Zadeh_Mihalcea_Morency_Hoque_2021}
Md~Kamrul Hasan, Sangwu Lee, Wasifur Rahman, Amir Zadeh, Rada Mihalcea,
  Louis-Philippe Morency, and Ehsan Hoque. 2021.
\newblock \href {https://ojs.aaai.org/index.php/AAAI/article/view/17534} {Humor
  knowledge enriched transformer for understanding multimodal humor}.
\newblock \emph{Proceedings of the AAAI Conference on Artificial Intelligence},
  35(14):12972--12980.

\bibitem[{Ivanko and Pexman(2003)}]{ivanko2003context}
Stacey~L. Ivanko and Penny~M. Pexman. 2003.
\newblock \href {https://doi.org/10.1207/S15326950DP3503\_2} {Context
  incongruity and irony processing}.
\newblock \emph{Discourse Processes}, 35(3):241--279.

\bibitem[{Joshi et~al.(2017)Joshi, Bhattacharyya, and
  Carman}]{1joshi2017automatic}
Aditya Joshi, Pushpak Bhattacharyya, and Mark~J. Carman. 2017.
\newblock \href {https://doi.org/10.1145/3124420} {Automatic sarcasm detection:
  A survey}.
\newblock \emph{ACM Comput. Surv.}, 50(5).

\bibitem[{Joshi et~al.(2015)Joshi, Sharma, and
  Bhattacharyya}]{joshi:sarcsam:incongruity:2015}
Aditya Joshi, Vinita Sharma, and Pushpak Bhattacharyya. 2015.
\newblock \href {https://doi.org/10.3115/v1/P15-2124} {Harnessing context
  incongruity for sarcasm detection}.
\newblock In \emph{Proceedings of the 53rd Annual Meeting of the Association
  for Computational Linguistics and the 7th International Joint Conference on
  Natural Language Processing (Volume 2: Short Papers)}, pages 757--762,
  Beijing, China. Association for Computational Linguistics.

\bibitem[{Kay et~al.(2017)Kay, Carreira, Simonyan, Zhang, Hillier,
  Vijayanarasimhan, Viola, Green, Back, Natsev, Suleyman, and
  Zisserman}]{kay2017kinetics}
Will Kay, Joao Carreira, Karen Simonyan, Brian Zhang, Chloe Hillier, Sudheendra
  Vijayanarasimhan, Fabio Viola, Tim Green, Trevor Back, Paul Natsev, Mustafa
  Suleyman, and Andrew Zisserman. 2017.
\newblock \href {http://arxiv.org/abs/1705.06950} {The kinetics human action
  video dataset}.

\bibitem[{Kreuz and Caucci(2007)}]{kreuz-caucci:2007:sarcasm:lexical}
Roger Kreuz and Gina Caucci. 2007.
\newblock \href {https://aclanthology.org/W07-0101} {Lexical influences on the
  perception of sarcasm}.
\newblock In \emph{Proceedings of the Workshop on Computational Approaches to
  Figurative Language}, pages 1--4, Rochester, New York. Association for
  Computational Linguistics.

\bibitem[{Lewis et~al.(2020)Lewis, Liu, Goyal, Ghazvininejad, Mohamed, Levy,
  Stoyanov, and Zettlemoyer}]{lewis2019bart}
Mike Lewis, Yinhan Liu, Naman Goyal, Marjan Ghazvininejad, Abdelrahman Mohamed,
  Omer Levy, Veselin Stoyanov, and Luke Zettlemoyer. 2020.
\newblock \href {https://doi.org/10.18653/v1/2020.acl-main.703} {{BART}:
  Denoising sequence-to-sequence pre-training for natural language generation,
  translation, and comprehension}.
\newblock In \emph{Proceedings of the 58th Annual Meeting of the Association
  for Computational Linguistics}, pages 7871--7880, Online. Association for
  Computational Linguistics.

\bibitem[{Lin(2004)}]{lin-2004-rouge}
Chin-Yew Lin. 2004.
\newblock \href {https://aclanthology.org/W04-1013} {{ROUGE}: A package for
  automatic evaluation of summaries}.
\newblock In \emph{Text Summarization Branches Out}, pages 74--81, Barcelona,
  Spain. Association for Computational Linguistics.

\bibitem[{Liu et~al.(2020)Liu, Gu, Goyal, Li, Edunov, Ghazvininejad, Lewis, and
  Zettlemoyer}]{liu2020multilingual}
Yinhan Liu, Jiatao Gu, Naman Goyal, Xian Li, Sergey Edunov, Marjan
  Ghazvininejad, Mike Lewis, and Luke Zettlemoyer. 2020.
\newblock \href {https://doi.org/10.1162/tacl_a_00343} {Multilingual denoising
  pre-training for neural machine translation}.
\newblock \emph{Transactions of the Association for Computational Linguistics},
  8:726--742.

\bibitem[{Mishra et~al.(2019)Mishra, Tater, and
  Sankaranarayanan}]{mishra-etal-2019-modular}
Abhijit Mishra, Tarun Tater, and Karthik Sankaranarayanan. 2019.
\newblock \href {https://doi.org/10.18653/v1/D19-1636} {A modular architecture
  for unsupervised sarcasm generation}.
\newblock In \emph{Proceedings of the 2019 Conference on Empirical Methods in
  Natural Language Processing and the 9th International Joint Conference on
  Natural Language Processing (EMNLP-IJCNLP)}, pages 6144--6154, Hong Kong,
  China. Association for Computational Linguistics.

\bibitem[{Olkoniemi et~al.(2016)Olkoniemi, Ranta, and
  Kaakinen}]{olkoniemi2016individual}
Henri Olkoniemi, Henri Ranta, and Johanna~K Kaakinen. 2016.
\newblock \href
  {https://doi.org/https://psycnet.apa.org/doi/10.1037/xlm0000176} {Individual
  differences in the processing of written sarcasm and metaphor: Evidence from
  eye movements.}
\newblock \emph{Journal of Experimental Psychology: Learning, Memory, and
  Cognition}, 42(3):433.

\bibitem[{Oraby et~al.(2017)Oraby, Harrison, Misra, Riloff, and
  Walker}]{oraby2017serious}
Shereen Oraby, Vrindavan Harrison, Amita Misra, Ellen Riloff, and Marilyn
  Walker. 2017.
\newblock \href {https://doi.org/10.18653/v1/W17-5537} {Are you serious?:
  Rhetorical questions and sarcasm in social media dialog}.
\newblock In \emph{Proceedings of the 18th Annual {SIG}dial Meeting on
  Discourse and Dialogue}, pages 310--319, Saarbr{\"u}cken, Germany.
  Association for Computational Linguistics.

\bibitem[{Ortega-Bueno et~al.(2019)Ortega-Bueno, Rangel,
  Hern{\'a}ndez~Far{\i}as, Rosso, Montes-y G{\'o}mez, and
  Medina~Pagola}]{ortega2019overview}
Reynier Ortega-Bueno, Francisco Rangel, D~Hern{\'a}ndez~Far{\i}as, Paolo Rosso,
  Manuel Montes-y G{\'o}mez, and Jos{\'e}~E Medina~Pagola. 2019.
\newblock \href {http://ceur-ws.org/Vol-2421/IroSvA_overview.pdf} {Overview of
  the task on irony detection in spanish variants}.
\newblock In \emph{Proceedings of the Iberian languages evaluation forum
  (IberLEF 2019), co-located with 34th conference of the Spanish Society for
  natural language processing (SEPLN 2019). CEUR-WS. org}, volume 2421, pages
  229--256.

\bibitem[{Pan et~al.(2020)Pan, Lin, Fu, Qi, and Wang}]{pan-etal-2020-modeling}
Hongliang Pan, Zheng Lin, Peng Fu, Yatao Qi, and Weiping Wang. 2020.
\newblock \href {https://doi.org/10.18653/v1/2020.findings-emnlp.124} {Modeling
  intra and inter-modality incongruity for multi-modal sarcasm detection}.
\newblock In \emph{Findings of the Association for Computational Linguistics:
  EMNLP 2020}, pages 1383--1392, Online. Association for Computational
  Linguistics.

\bibitem[{Papineni et~al.(2002)Papineni, Roukos, Ward, and
  Zhu}]{papineni2002bleu}
Kishore Papineni, Salim Roukos, Todd Ward, and Wei-Jing Zhu. 2002.
\newblock \href {https://doi.org/10.3115/1073083.1073135} {{B}leu: a method for
  automatic evaluation of machine translation}.
\newblock In \emph{Proceedings of the 40th Annual Meeting of the Association
  for Computational Linguistics}, pages 311--318, Philadelphia, Pennsylvania,
  USA. Association for Computational Linguistics.

\bibitem[{Peled and Reichart(2017)}]{peled-reichart-2017-sarcasm}
Lotem Peled and Roi Reichart. 2017.
\newblock \href {https://doi.org/10.18653/v1/P17-1155} {Sarcasm {SIGN}:
  Interpreting sarcasm with sentiment based monolingual machine translation}.
\newblock In \emph{Proceedings of the 55th Annual Meeting of the Association
  for Computational Linguistics (Volume 1: Long Papers)}, pages 1690--1700,
  Vancouver, Canada. Association for Computational Linguistics.

\bibitem[{Rajpurkar et~al.(2016)Rajpurkar, Zhang, Lopyrev, and
  Liang}]{rajpurkar2016squad}
Pranav Rajpurkar, Jian Zhang, Konstantin Lopyrev, and Percy Liang. 2016.
\newblock Squad: 100,000+ questions for machine comprehension of text.
\newblock \emph{arXiv preprint arXiv:1606.05250}.

\bibitem[{Roberts and Kreuz(1994)}]{roberts1994people}
Richard~M. Roberts and Roger~J. Kreuz. 1994.
\newblock \href {https://doi.org/10.1111/j.1467-9280.1994.tb00653.x} {Why do
  people use figurative language?}
\newblock \emph{Psychological Science}, 5(3):159--163.

\bibitem[{Rockwell(2007)}]{rockwell2007vocal}
Patricia Rockwell. 2007.
\newblock \href {https://doi.org/https://doi.org/10.1007/s10936-006-9049-0}
  {Vocal features of conversational sarcasm: A comparison of methods}.
\newblock \emph{Journal of psycholinguistic research}, 36(5):361--369.

\bibitem[{See et~al.(2017)See, Liu, and Manning}]{see2017get}
Abigail See, Peter~J. Liu, and Christopher~D. Manning. 2017.
\newblock \href {https://doi.org/10.18653/v1/P17-1099} {Get to the point:
  Summarization with pointer-generator networks}.
\newblock In \emph{Proceedings of the 55th Annual Meeting of the Association
  for Computational Linguistics (Volume 1: Long Papers)}, pages 1073--1083,
  Vancouver, Canada. Association for Computational Linguistics.

\bibitem[{Srivastava et~al.(2020)Srivastava, Varshney, Kumari, and
  Srivastava}]{srivastava-etal-2020-novel}
Himani Srivastava, Vaibhav Varshney, Surabhi Kumari, and Saurabh Srivastava.
  2020.
\newblock \href {https://doi.org/10.18653/v1/2020.figlang-1.14} {A novel
  hierarchical {BERT} architecture for sarcasm detection}.
\newblock In \emph{Proceedings of the Second Workshop on Figurative Language
  Processing}, pages 93--97, Online. Association for Computational Linguistics.

\bibitem[{Swami et~al.(2018)Swami, Khandelwal, Singh, Akhtar, and
  Shrivastava}]{swami2018corpus}
Sahil Swami, Ankush Khandelwal, Vinay Singh, Syed~Sarfaraz Akhtar, and Manish
  Shrivastava. 2018.
\newblock A corpus of english-hindi code-mixed tweets for sarcasm detection.
\newblock \emph{arXiv preprint arXiv:1805.11869}.

\bibitem[{Tabacaru and Lemmens(2014)}]{Tabacaru_Lemmens_2014}
Sabina Tabacaru and Maarten Lemmens. 2014.
\newblock \href {https://doi.org/10.7592/EJHR2014.2.2.tabacaru} {Raised
  eyebrows as gestural triggers in humour: The case of sarcasm and
  hyper-understanding}.
\newblock \emph{The European Journal of Humour Research}, 2(2):11–31.

\bibitem[{Tay et~al.(2018)Tay, Luu, Hui, and Su}]{tay-etal-2018-reasoning}
Yi~Tay, Anh~Tuan Luu, Siu~Cheung Hui, and Jian Su. 2018.
\newblock \href {https://doi.org/10.18653/v1/P18-1093} {Reasoning with sarcasm
  by reading in-between}.
\newblock In \emph{Proceedings of the 56th Annual Meeting of the Association
  for Computational Linguistics (Volume 1: Long Papers)}, pages 1010--1020,
  Melbourne, Australia. Association for Computational Linguistics.

\bibitem[{Tsur et~al.(2010)Tsur, Davidov, and Rappoport}]{tsur:sarcasm:2010}
Oren Tsur, Dmitry Davidov, and Ari Rappoport. 2010.
\newblock \href {https://ojs.aaai.org/index.php/ICWSM/article/view/14018}
  {Icwsm — a great catchy name: Semi-supervised recognition of sarcastic
  sentences in online product reviews}.
\newblock \emph{Proceedings of the International AAAI Conference on Web and
  Social Media}, 4(1):162--169.

\bibitem[{Vaswani et~al.(2017)Vaswani, Shazeer, Parmar, Uszkoreit, Jones,
  Gomez, Kaiser, and Polosukhin}]{vaswani2017attention}
Ashish Vaswani, Noam Shazeer, Niki Parmar, Jakob Uszkoreit, Llion Jones,
  Aidan~N Gomez, \L~ukasz Kaiser, and Illia Polosukhin. 2017.
\newblock \href
  {https://proceedings.neurips.cc/paper/2017/file/3f5ee243547dee91fbd053c1c4a845aa-Paper.pdf}
  {Attention is all you need}.
\newblock In \emph{Advances in Neural Information Processing Systems},
  volume~30. Curran Associates, Inc.

\bibitem[{Wang et~al.(2018)Wang, Singh, Michael, Hill, Levy, and
  Bowman}]{wang2018glue}
Alex Wang, Amanpreet Singh, Julian Michael, Felix Hill, Omer Levy, and Samuel~R
  Bowman. 2018.
\newblock Glue: A multi-task benchmark and analysis platform for natural
  language understanding.
\newblock \emph{arXiv preprint arXiv:1804.07461}.

\bibitem[{Wellman(2014)}]{wellman2014making}
Henry~M Wellman. 2014.
\newblock \emph{Making minds: How theory of mind develops}.
\newblock Oxford University Press.

\bibitem[{Xiong et~al.(2019)Xiong, Zhang, Zhu, and Yang}]{xiong2019sarcasm}
Tao Xiong, Peiran Zhang, Hongbo Zhu, and Yihui Yang. 2019.
\newblock \href {https://doi.org/10.1145/3308558.3313735} {Sarcasm detection
  with self-matching networks and low-rank bilinear pooling}.
\newblock In \emph{The World Wide Web Conference}, WWW '19, page 2115–2124,
  New York, NY, USA. Association for Computing Machinery.

\bibitem[{Xu et~al.(2020)Xu, Zeng, and Mao}]{xu-etal-2020-reasoning}
Nan Xu, Zhixiong Zeng, and Wenji Mao. 2020.
\newblock \href {https://doi.org/10.18653/v1/2020.acl-main.349} {Reasoning with
  multimodal sarcastic tweets via modeling cross-modality contrast and semantic
  association}.
\newblock In \emph{Proceedings of the 58th Annual Meeting of the Association
  for Computational Linguistics}, pages 3777--3786, Online. Association for
  Computational Linguistics.

\bibitem[{Yang et~al.(2019)Yang, Li, Wong, Chao, Wang, and
  Tu}]{Yang_Li_Wong_Chao_Wang_Tu_2019}
Baosong Yang, Jian Li, Derek~F. Wong, Lidia~S. Chao, Xing Wang, and Zhaopeng
  Tu. 2019.
\newblock \href {https://doi.org/10.1609/aaai.v33i01.3301387} {Context-aware
  self-attention networks}.
\newblock \emph{Proceedings of the AAAI Conference on Artificial Intelligence},
  33(01):387--394.

\bibitem[{Zhang et~al.(2019)Zhang, Kishore, Wu, Weinberger, and
  Artzi}]{zhang2019bertscore}
Tianyi Zhang, Varsha Kishore, Felix Wu, Kilian~Q Weinberger, and Yoav Artzi.
  2019.
\newblock \href {https://openreview.net/forum?id=SkeHuCVFDr} {Bertscore:
  Evaluating text generation with bert}.
\newblock In \emph{International Conference on Learning Representations}.

\end{thebibliography}
\bibliographystyle{acl_natbib}

\newpage
\clearpage
\appendix
\section{Appendix}
\subsection{Embedding Space for BART and mBART}
\label{sec:bart_vs_mbart}
\begin{figure}[ht]
\subfloat[Pre-trained]{
\includegraphics[width=\columnwidth]{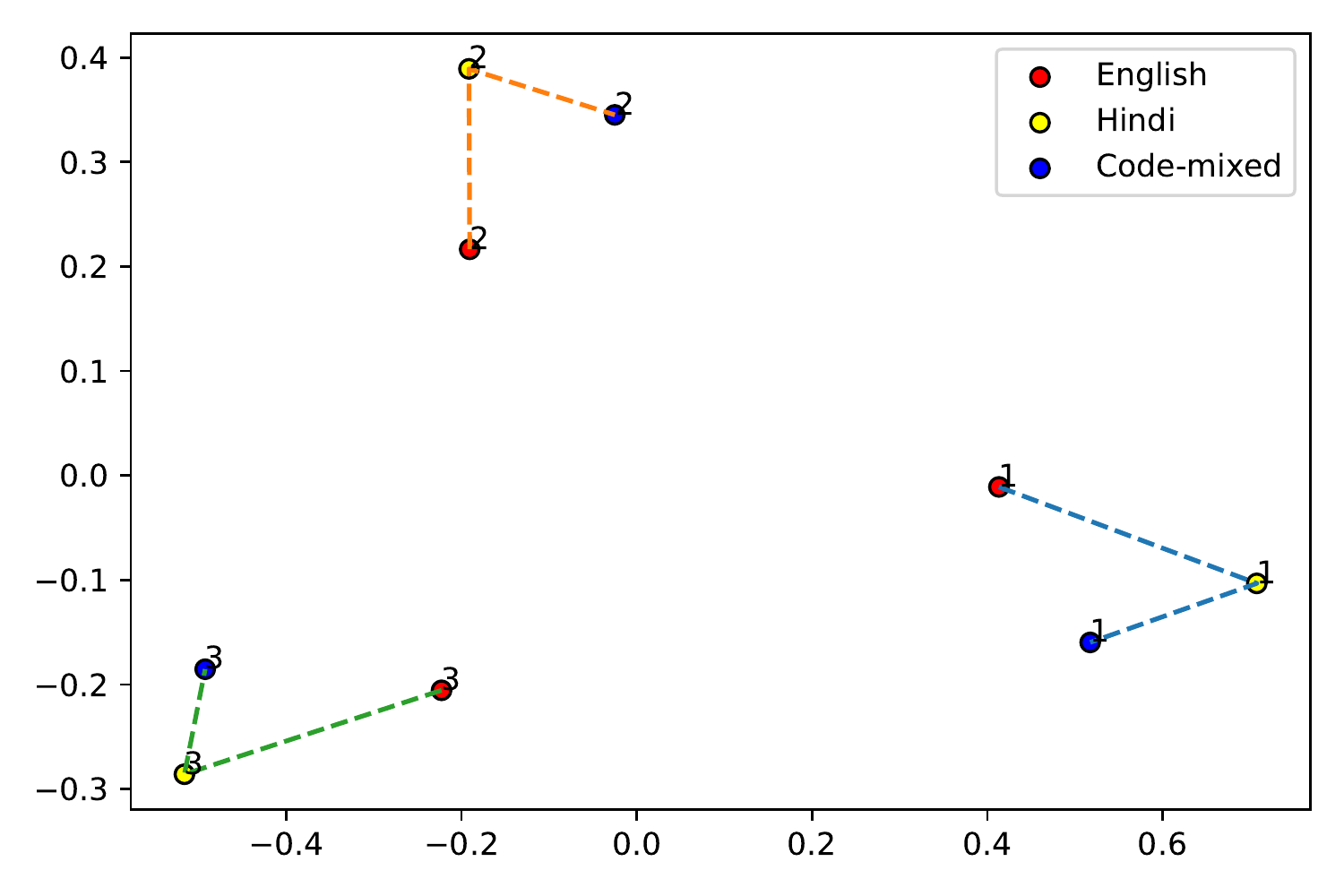}
}\\
\subfloat[Fine-tuned]{
\includegraphics[width=\columnwidth]{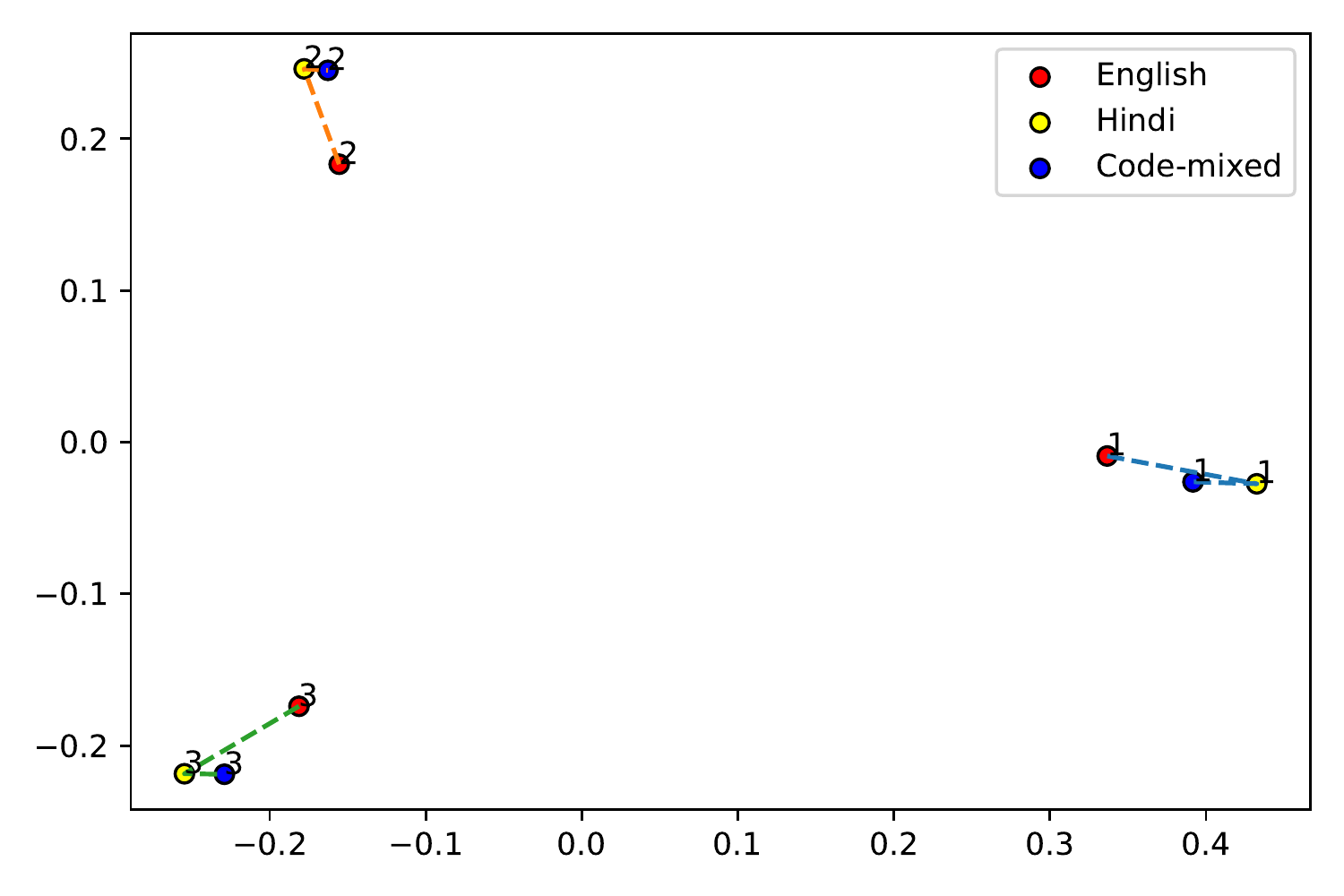}
}
\caption{Embedding space for BART before and after fine-tuning on sarcasm explanation in dialogues.}
\label{fig:embed_space}
\end{figure}

We compared various text based unimodal methods for our task. Although BART is performing the best for SED, it is important to note that BART is pre-trained on English datasets (GLUE \citep{wang2018glue} and SQUAD \citep{rajpurkar2016squad}). In order to explore how the representation learning is being transferred to a code-mixed setting, we analyse the embedding space learnt by the model before and after fine-tuning it for our task. We considered three random utterances from \dataset\ and created three copies of them- one in English, one in Hindi (romanised), and one without modification i.e. code-mixed. Figure \ref{fig:embed_space} illustrates the PCA plot for the embeddings obtained for these nine utterance representations obtained by BART before and after fine-tuning on our task. It is interesting to note that even before any fine-tuning the Hindi, English, and code-mixed representations lie closer to each other and they shift further closer when we fine-tune our model. This phenomenon can be justified as out input is of romanised code-mixed format and thus we can assume that representations are already being captured by the pre-trained model. Fine-tuning helps us understand the Hindi part of the input. Table \ref{tab:embed_cosine} shows the cosine distance between the representations obtained for English-Hindi, English-Code mixed, and Code mixed-Hindi utterances for the sample utterances. It can be clearly seen that the distance is decreasing after fine-tuning.

\begin{table}[h!]
\centering
\resizebox{\columnwidth}{!}{%
\begin{tabular}{|l|cc|cc|cc|}
\hline
\multirow{2}{*}{\textbf{Example}} & \multicolumn{2}{c|}{\textbf{English-Hindi}} & \multicolumn{2}{c|}{\textbf{English-Code mixed}} & \multicolumn{2}{c|}{\textbf{Code mixed-Hindi}} \\ \cline{2-7} 
 & \multicolumn{1}{c|}{\textbf{PT}} & \textbf{FT} & \multicolumn{1}{c|}{\textbf{PT}} & \textbf{FT} & \multicolumn{1}{c|}{\textbf{PT}} & \textbf{FT} \\ \hline
\textbf{1} & \multicolumn{1}{c|}{0.183} & \bf 0.067 & \multicolumn{1}{c|}{0.014} & \bf 0.006 & \multicolumn{1}{c|}{0.118} & \bf 0.056 \\ \hline
\textbf{2} & \multicolumn{1}{c|}{0.282} & \bf 0.093 & \multicolumn{1}{c|}{0.017} & \bf 0.007 & \multicolumn{1}{c|}{0.197} & \bf 0.066 \\ \hline
\textbf{3} & \multicolumn{1}{c|}{0.321} & \bf 0.113 & \multicolumn{1}{c|}{0.065} & \bf 0.020 & \multicolumn{1}{c|}{0.132} & \bf 0.057 \\ \hline
\end{tabular}%
}
\caption{Cosine distance between three random samples from the dataset before and after fine-tuning. (PT: pre-trained; FT: fine-tuned)}
\label{tab:embed_cosine}
\end{table}

\subsection{Fusion at Different Layers}
\label{sec:app_fusion}
We fuse the multimodal information of audio and video in the BART encoder using the proposed fusion mechanism before different layers of the BART encoder. Table \ref{tab:ablation_fusion} shows the results we obtain when the fusion happens at different layers. We obtain the best results when the fusion happens before layer $6$ i.e. the last layer of the encoder. This can be attributed to the fact that since there is only one layer of encoder after the fusion, the multimodal information is being retained efficiently and thus being decoded more accurately.
\begin{table}[h!]
\centering
\resizebox{0.8\columnwidth}{!}{%
\begin{tabular}{|c|c|c|c|}
\hline
\textbf{Fusion before layer \#} & \textbf{R1} & \textbf{R2} & \textbf{RL} \\ \hline
\textbf{1} & 37.27 & 13.95 & 35.24 \\ \hline
\textbf{2} & 37.63 & 14.32 & 35.57 \\ \hline
\textbf{3} & 36.73 & 13.15 & 34.63 \\ \hline
\textbf{4} & 37.61 & 14.98 & 36.04 \\ \hline
\textbf{5} & 37.34 & 13.67 & 35.48 \\ \hline
\textbf{6} & \textbf{39.69} & \textbf{17.10} & \textbf{37.37} \\ \hline
\end{tabular}%
}
\caption{ROUGE scores for fusion before different layers (R1/2/L: ROUGE1/2/L).}
\label{tab:ablation_fusion}
\end{table}

\subsection{More Qualitative Analysis}
\label{sec:app_qual_analysis}
Table \ref{tab:bart_vs_mbart} highlights one of many cases where BART is able to capture the essence of sarcasm in a better way when compared to mBART. While mBART gives us an incorrect and incoherent explanation, BART generates an explanation which essentially means the same as the ground truth explanation. The inclusion of audio modality in the unimodal system often helps in generating preferable explanations, as shown in Table \ref{tab:audio_helps}. AVII-TA is able to capture the essense of sarcasm in the dialogue while the unimodal systems were not able to do so. Furthermore, video modality facilitates even better understanding of sarcasm as illustrated in Table \ref{tab:video_helps}. AVII-TV is able to generate the best results while audio may act as noise in this particular example.
\begin{table}[h!]
    \centering
    \resizebox{0.9\columnwidth}{!}{%
    \begin{tabular}{p{5em}p{20em}}
    \hline \multicolumn{2}{l}{{\begin{tabular}[c]{@{}p{27em}@{}}MAYA: Sahil, beta tum bhi soche ho ki maine Monisha ki speech churai? \textit{\color{blue} Sahil, do you also think that I stole Monisha's speech?}\\
    INDRAVARDHAN: Haan.\textit{\color{blue}Yes.}\\
    MAYA: Are darling maine to speech ko chua bhi nahin. chhoti to germs nahin lag jaate? Kyunki Monisha ne mithaai box ki wrapper per likhi thi apni speech hath mein uthati to makkhiya bhanbhana ne lagti. \textit{\color{blue}Darling, I didn't even touch the speech. Would I not have got germs by touching it? Monisha used sweets wrapper to write her speech, if I would have picked it up, there would've been flies buzzing around me.}
    \end{tabular}}}\\
    \hline
    \rowcolor{LightGreen} \bf Gold & Maya ne Monisha ke speech ka mazak udaya. \textit{\color{blue}Maya makes fun of Monisha's speech.} \\
    \bf mBART & Maya kehti hai ki Monisha ka mazak udata hai
    \textit{\color{blue}Maya says that make fun of Monisha.}\\
    \rowcolor{LightCyan}\bf BART & Maya monisha ke speech ka mazak udati hai
    \textit{\color{blue}Maya makes fun of Monisha's speech.}\\
    \bf \modelTA & Maya monisha ke speech ka mazaak udati hai
    \textit{\color{blue}Maya says that make fun of Monisha.}\\
    \bf \modelTV & Maya mocks monisha kyunki wo rhe theek hai
    \textit{\color{blue}Maya mocks Monisha because she is okay.}\\
    \bf \modelTAV & Maya kehti hai ki uske speech bure hai
    \textit{\color{blue}Maya says that she didn't like the speech.}\\
    \hline
    \end{tabular}}
    \caption{BART v/s mBART: An example where explanation generated by BART is better than mBART.}
    \label{tab:bart_vs_mbart}
\end{table}

\begin{table}[h!]
    \centering
    \resizebox{0.9\columnwidth}{!}{%
    \begin{tabular}{p{5em}p{20em}}
    \hline
    \multicolumn{2}{l}{\begin{tabular}[c]{@{}p{27em}@{}}SAHIL: Ek minute, kya hai maa ji, humaare naatak mein ek bhi stree patra nahi hai, sare ladke hai. \textit{\color{blue}One minute, what is it ma'am, we don't have any female parts in our play, all are male}\\PRABHAVATI: To uss mein bhi kaunsi badi baat hai, mai ladka ban jaungi. Mere paas pant shirt to hai, moonche aapki de dena! \textit{\color{blue}So what is the big deal in it, I'll play a male. I have pant shirt, you give me your mustache.}\\INDRAVARDHAN: Cancel! Naatak cancel! Maa ji huaa aisa ki humaari jo bahu hai, uska ek chota sa accident ho gaya, to iss liye natak cancel! Monisha le jaao inhe. \textit{\color{blue}Cancel! Play cancel! Ma'am, what happened is, that our daughter in law had a small accident, that is why the play is cancelled. Monisha take her.}\\SAHIL: Aur aate aate apna ek chota sa accident bhi kara ke aao! \textit{\color{blue}And when you come, have a small accident too!}\end{tabular}} \\ \hline
    \rowcolor{LightGreen} \bf Gold & Sahil Monisha pe gussa hai as usne Prabhavati as an actress le aya.
    \textit{\color{blue}Sahil is angry on Monisha that she hired Prabhavati as an actress.} \\
    \bf mBART & Sahil ko Prabhavati ko role offer karne par taunt maarta hai.
    \textit{\color{blue}Sahil taunts because the role is being offered to Prabhavati.}\\
    \bf BART & Indravardhan Monisha ko taunt maarta hai ki uska ek chota sa accident bhi kara ke aao.
    \textit{\color{blue}Indravardhan taunts Monisha that she should have a small accident.}\\
    \rowcolor{LightCyan}\bf \modelTA & Sahil ko Prabhavati ko role offer nahi karna.
    \textit{\color{blue}Sahil does not want Prabhavati to have this role.}\\
    \bf \modelTV & Sahil Indravardhan ko ek accident keh ke uska majaak udaata hai.
    \textit{\color{blue}Calls Indravardhan an accident and makes fun of him.}\\
    \bf \modelTAV & Sahil ko Prabhavati ko role offer nahi karna.
    \textit{\color{blue}Sahil does not want Prabhavati to have this role.}\\
    \hline
    \end{tabular}}
    \caption{Audio helps: An example where audio modality helps in generating more fitting explanation.}
    \label{tab:audio_helps}
\end{table}

\begin{table}[h!]
    \centering
    \resizebox{0.9\columnwidth}{!}{%
    \begin{tabular}{p{5em}p{20em}}
    \hline
    \multicolumn{2}{l}{\begin{tabular}[c]{@{}p{27em}@{}}MAYA: Kshama? You mean Sahil Kshama ko pyaar karta hai!? \textit{\color{blue}Kshama? You mean Sahil loves Kshama?}\\SAHIL: Nahi, nahi! Ek minute, ek minute, mai kshama chahata hu. \textit{\color{blue}No no, One minute, one minute, I want forgiveness (kshama in hindi).}\\INDRAVARDHAN: Dekha, Kshama chahata hai! Chahata ka matlab pyaar karna hi hua na!? \textit{\color{blue}See, wants forgiveness! Wants means love only, no!?}\end{tabular}} \\ \hline
    \rowcolor{LightGreen} \bf Gold & Indravardhan Sahil ko tease karta hai ki vo Kshama se pyaar karta hai.\textit{\color{blue}Indravardhan teases Sahil by implying that he loves kshama (name of a girl in hindi meaning forgiveness)} \\
    \bf mBART & Indravardhan implies ki Sahil ek kshama chahata hai. \textit{\color{blue}Indravardhan implies that Sahil wants forgiveness.}\\
    \bf BART & Maya ko kshama chahata hai
    \textit{\color{blue}Maya wants forgiveness.}\\
    \bf \modelTA & Indravardhan Kshama ko pyaar karne par taunt maarta hai.
    \textit{\color{blue}Indravardhan taunts that he loves Kshama.}\\
    \rowcolor{LightCyan} \bf \modelTV & Indravardhan majaak mein kehta hai ki Sahil Kshama ko pyaar karta hai.
    \textit{\color{blue}Indravardhan jokes that Sahil loves Kshama}\\
    \bf \modelTAV & Indravardhan Rosesh ko Kshama ki matlab pyaar karne par taunt maarta hai.
    \textit{\color{blue}Indravardhan taunts Rosesh for loving the meaning of forgiveness.}\\
    \hline
    \end{tabular}}
    \caption{Video helps: An example where video modality helps in generating more fitting explanation.}
    \label{tab:video_helps}
\end{table}

\begin{table}[h!]
    \centering
    \resizebox{0.9\columnwidth}{!}{%
    \begin{tabular}{p{5em}p{20em}}
    \hline
    \multicolumn{2}{l}{\begin{tabular}[c]{@{}p{27em}@{}}MAYA: And this time I thought lets have a theme party! animals! Hum log sab animals banenge! \textit{\color{blue}And this time I thought lets have a theme party! animals! We will all be animals!}\\MONISHA: Walnut brownie bhi. \textit{\color{blue}And walnut brownie too.}\\MAYA: Mai hiran, Sahil horse, and Monisha chhipakalee! \textit{\color{blue}I'll be a deer, Sahil horse, and Monisha lizard!}\end{tabular}} \\\hline
    \rowcolor{LightGreen} \bf Gold & Maya Monisha ko chhipakalee keha kar uska majaak udaati hai.\textit{\color{blue}Maya makes fun of Monisha by comparing her with a lizard.} \\
    \bf mBART & Maya Monisha ko taunt maarti hai ki use animal themed party
    \textit{\color{blue}Maya taunts Monisha for her animal themed party.}\\
    \bf BART & Maya Monisha ko taunt maarti hai.
    \textit{\color{blue}Maya taunts Monisha.}\\
    \bf \modelTA & Maya implies ki vo animal mein theme party ke baare mein nahi banenge.
    \textit{\color{blue}Maya implies that she won't be in regarding animal themed party.}\\
    \bf \modelTV & Maya Monisha ke animal ke behaviour par taunt maarti hai.
    \textit{\color{blue}Maya taunts Monisha for her animal behaviour.}\\
    \rowcolor{LightCyan} \bf \modelTAV & Maya Monisha ko animal kaha ke taunt maarti hai.
    \textit{\color{blue}Maya taunts Monisha by calling her an animal.}\\
    \hline
    \end{tabular}}
    \caption{Audio and video helps: An example where audio and video modality together helps in generating better explanation.}
    \label{tab:audio_video_helps}
\end{table}

\end{document}